\documentclass[lettersize,journal,twoside]{IEEEtran}
\usepackage{amsmath,amsfonts}
\usepackage{algorithmic}
\usepackage{algorithm}
\usepackage{array}
\usepackage[caption=false,font=normalsize,labelfont=sf,textfont=sf]{subfig}
\usepackage{textcomp} 
\usepackage{stfloats}
\usepackage{url}
\usepackage{verbatim}
\usepackage{graphicx}
\usepackage{cite}
\usepackage{wrapfig}
\usepackage[table]{xcolor}
\usepackage{threeparttable}

\usepackage{amsmath,amssymb,amsfonts}
\usepackage{textcomp}
\usepackage{tikz}
\usepackage{hyperref}
\usepackage{booktabs}
\usepackage{multirow}
\usepackage{tcolorbox}
\usepackage{soul}
\usepackage{adjustbox}
\usepackage{diagbox}
\usepackage{enumitem}
\usepackage{savefnmark}

\usepackage{pgfplots}
\pgfplotsset{compat=newest}

\usepackage{array}
\newcolumntype{P}[1]{>{\centering\arraybackslash}p{#1}}

\usepackage{orcidlink}

\DeclareMathOperator*{\argmax}{argmax} 
\DeclareMathOperator*{\metric}{metric} 
\DeclareMathOperator*{\softmax}{softmax}

\newcommand{\Hquad}{\hspace{0.5em}}

\DeclareMathAlphabet{\mathcal}{OMS}{cmsy}{m}{n}

\usepackage[labelfont=rm,textfont=rm,labelsep=period]{caption}   
\DeclareCaptionFormat{customTable}
{%
    \textsc{#1#2 #3}
}
\begin{document}

\title{TrICy: Trigger-guided Data-to-text Generation with Intent aware Attention-Copy}

\author {
    Vibhav Agarwal\textsuperscript{$\dagger$} \orcidlink{0000-0002-2029-9885},
    Sourav Ghosh\textsuperscript{$\dagger$} \orcidlink{0000-0003-1866-1408},
    Harichandana BSS  \orcidlink{0000-0002-6123-2249},
    Himanshu Arora \orcidlink{0009-0004-0199-6425},
    Barath Raj Kandur Raja  \orcidlink{0000-0003-0451-2452}
    \\
\thanks{$\dagger$ \textit{Sourav Ghosh and Vibhav Agarwal contributed equally to this work.}}
\thanks{
    Manuscript received 27 March 2023; revised 12 October 2023 and 08 December 2023; accepted 27 December 2023. Date of publication 12 January 2024. The associate editor coordinating the review of this manuscript and approving it for publication was Prof. Hung-yi Lee. \textit{(Corresponding authors: Sourav Ghosh, Vibhav Agarwal).}

    The authors are with Samsung R\&D Institute Bangalore, 560037 India (e-mail: vibhav.a@samsung.com; sourav.ghosh@samsung.com; hari.ss@ samsung.com; him.arora@samsung.com; barathraj.kr@samsung.com).

    Digital Object Identifer \href{https://doi.org/10.1109/TASLP.2024.3353574}{10.1109/TASLP.2024.3353574}
}}

\markboth{IEEE/ACM Transactions on Audio, Speech, and Language Processing,
2024}%
{Agarwal and Ghosh \MakeLowercase{\textit{et al.}}: {TrICy}: Trigger-guided Data-to-text Generation with Intent aware Attention-Copy}

\IEEEpubid{
    \begin{minipage}{\textwidth}\ \\[12pt]
    \centering
      2329-9290 \copyright\;2024 IEEE. Personal use is permitted, but republication/redistribution requires IEEE permission.\\ 
      See \url{https://www.ieee.org/publications/rights/index.html} for more information.
    \end{minipage}
}

\maketitle

\begin{abstract}
Data-to-text (D2T) generation is a crucial task in many natural language understanding (NLU) applications and forms the foundation of task-oriented dialog systems. In the context of conversational AI solutions that can work directly with local data on the user's device, architectures utilizing large pre-trained language models (PLMs) are impractical for on-device deployment due to a high memory footprint. To this end, we propose TrICy, a novel lightweight framework for an enhanced D2T task that generates text sequences based on the intent in context and may further be guided by user-provided triggers.
We leverage an attention-copy mechanism to predict out-of-vocabulary (OOV) words accurately.
Performance analyses on E2E NLG dataset \cite{novikova-etal-2017-e2e} (BLEU: 66.43\%, ROUGE-L: 70.14\%), WebNLG dataset \cite{gardent-etal-2017-creating} (BLEU: \textit{Seen} 64.08\%, \textit{Unseen} 52.35\%), and our Custom dataset related to text messaging applications, showcase our architecture's effectiveness. Moreover, we show that by leveraging an optional trigger input, data-to-text generation quality increases significantly and achieves the new SOTA score of 69.29\% BLEU for E2E NLG. 
Furthermore, our analyses show that TrICy achieves at least 24\% and 3\% improvement in BLEU and METEOR respectively over LLMs like GPT-3, ChatGPT, and Llama 2.
We also demonstrate that in some scenarios, performance improvement due to triggers is observed even when they are absent in training.\looseness=-1 

\end{abstract}

\begin{IEEEkeywords}
Data-to-text, Natural Language Generation, Edge devices.
\end{IEEEkeywords}

\section{Introduction}\label{sec:introduction}

\begin{figure}[t]
	\centering
	\includegraphics[width=\linewidth]{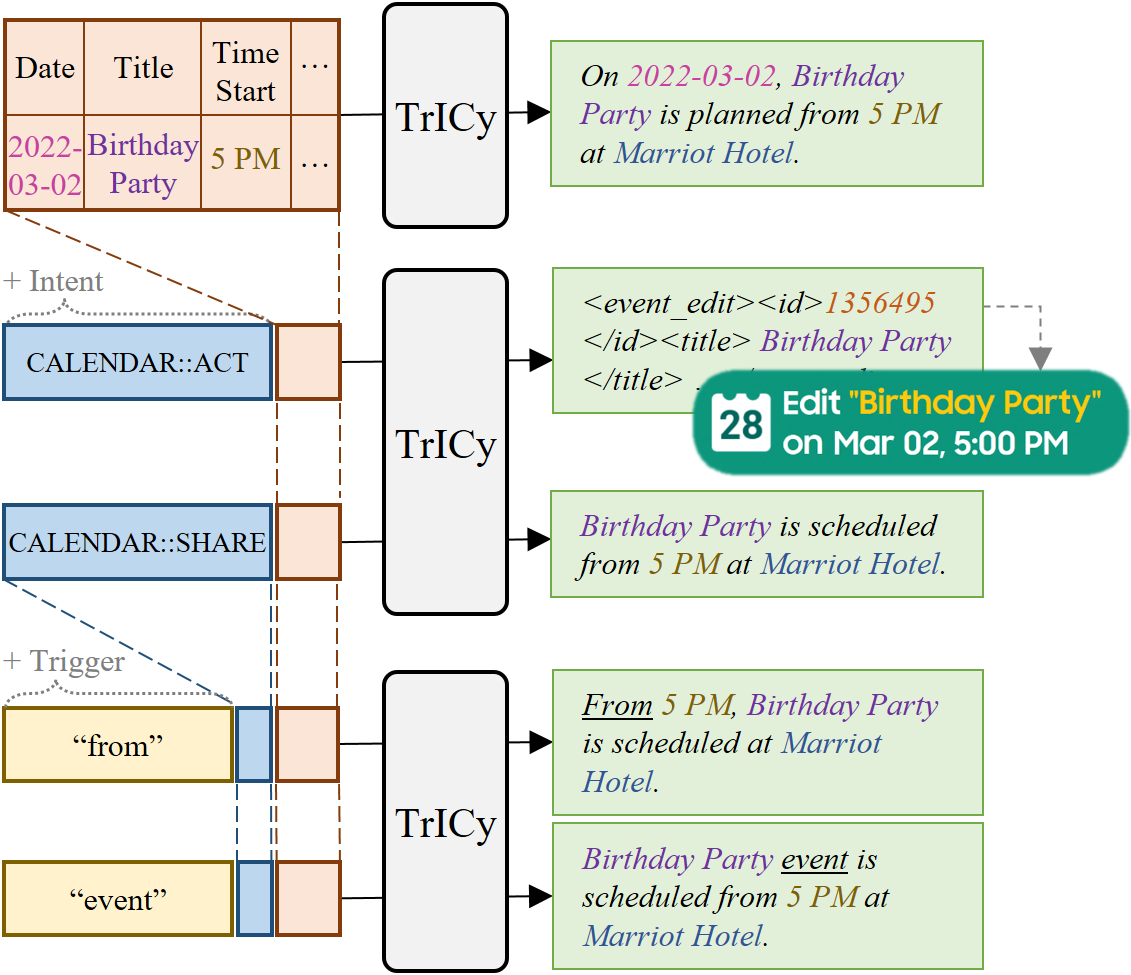}
	\caption{\textbf{Context-aware D2T generation:} {\normalfont based on \textit{user application data}, \textit{message intent}, and \textit{trigger text}. Generated text may include natural language responses or markup text for downstream use cases.}}
	\label{fig:newTeaser}
\end{figure}

Applications of Natural Language Generation (NLG) \cite{gatt2018survey, puduppully-etal-2019-data, ma-etal-2019-key, chen-etal-2019-enhancing} include tasks like text summarization, report generation, dialog response generation, etc. Data-to-text (D2T) in NLG deals with capturing a \textit{useful subset} of an instance of data in a generated syntactically correct text sequence. For example, given an input data unit \textit{\{~\texttt{name}: ``John'', \texttt{phone}: ``9876543210'', \texttt{email}: ``john@example.com'', \texttt{address}: ``404, Cyber Street''~\}}, the aim may be to generate a text description like \textit{``You can connect with John at 9876543210 or john@example.com''}.\looseness=-1

Large Language Models (LLMs) like GPT-4 have recently taken the NLP community by storm \cite{chatgpt2022, openai2023gpt4}. A prominent task solved by these is presenting of factual data in the form of (a) natural language responses, or (b) structured markup like tables \cite{new-bing}. However, LLMs are unsuitable for on-device integration due to large size (large number of parameters) and high latency. Moreover, a server-client deployment introduces the risk of exposing end user's private data.\looseness=-1

Contributions from recent NLP research have given an impetus to D2T generation task. Gardent et al. \cite{gardent-etal-2017-creating} proposed WebNLG task to generate text descriptions of the given input triplets describing facts. Around the same time, Novikova et al. \cite{novikova-etal-2017-e2e} proposed E2E task to form restaurant reviews according to the given restaurant meaning representations (MRs). Noticeably, many applications of D2T, like knowledge-based dialog systems, struggle with Out-of-Vocabulary (OOV) words that are frequent in input entities like Personally Identifiable Information (PII) or dates. Thus, these typical D2T generation tasks require fetching the relevant entities from input and adding them into the output sequence, referred to as ``copying mechanism'' \cite{gu-etal-2016-incorporating}.\looseness=-1 
\IEEEpubidadjcol

However, while the key challenge in D2T generation remains the surface realization of content, an important shortcoming of existing proposals arises from the fact that most approaches, ranging from simple template-based ones to complex neural networks, predict a single \textit{best} sequence from a given data unit. In real-world applications, it is often required to generate alternative text sequences based on the underlying intent behind the data retrieval and the immediate priorities of an end-user, as seen in Fig.~\ref{fig:newTeaser}.\looseness=-1

\begin{figure}[t]
	\centering
	\begin{minipage}{0.715\linewidth}
		\includegraphics[width=\linewidth]{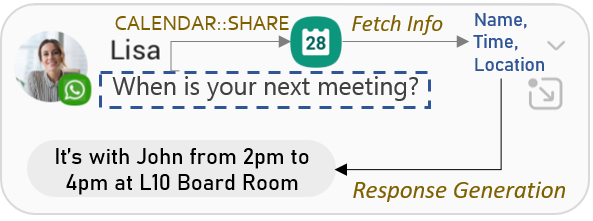}
		\caption*{\small (a) Sharing data}
		\label{fig:teaserPhone}
	\end{minipage}
	\begin{minipage}{0.27\linewidth}
		\includegraphics[width=\linewidth]{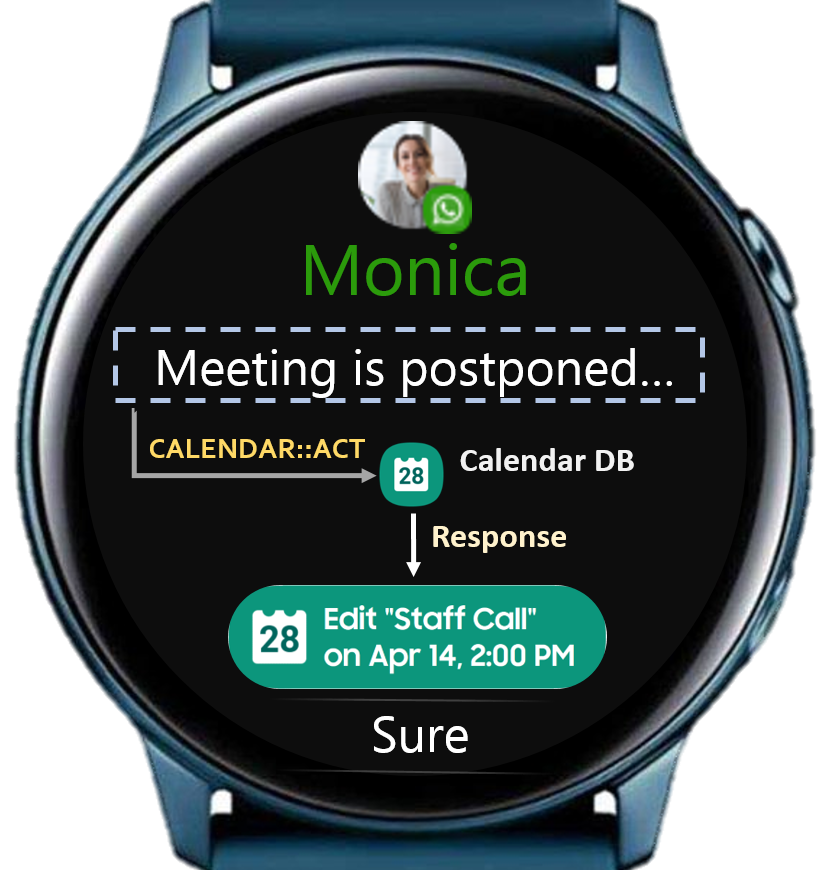}
		\caption*{\small (b) Acting on data}
		\label{fig:teaserWatch}
	\end{minipage}
	\caption{\textbf{Intent-aware response generation:} {\normalfont based on \textit{message intent} and \textit{user application data}}}
	\label{fig:teaser}
\end{figure}

\begin{figure}[t]
	\centering
	\begin{minipage}[t]{0.42\linewidth}
		\includegraphics[width=\linewidth, trim={0 1cm 0 6.8cm}, clip]{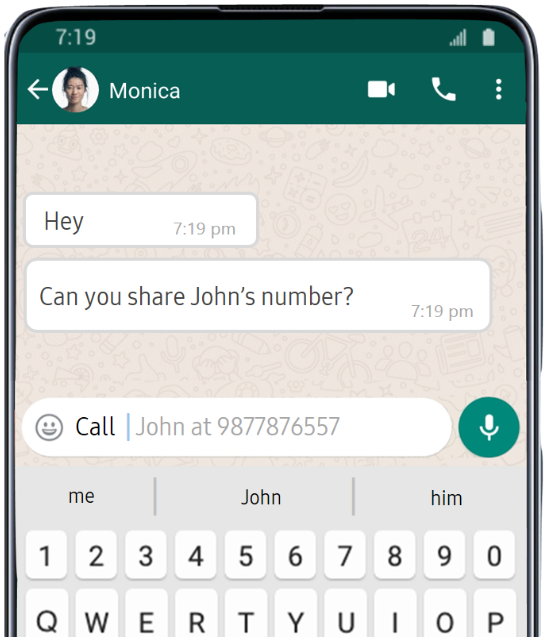}
		\label{fig:smartComposeSimple_a}
	\end{minipage}
	\begin{minipage}[t]{0.42\linewidth}
		\includegraphics[width=\linewidth, trim={0 1cm 0 6.8cm}, clip]{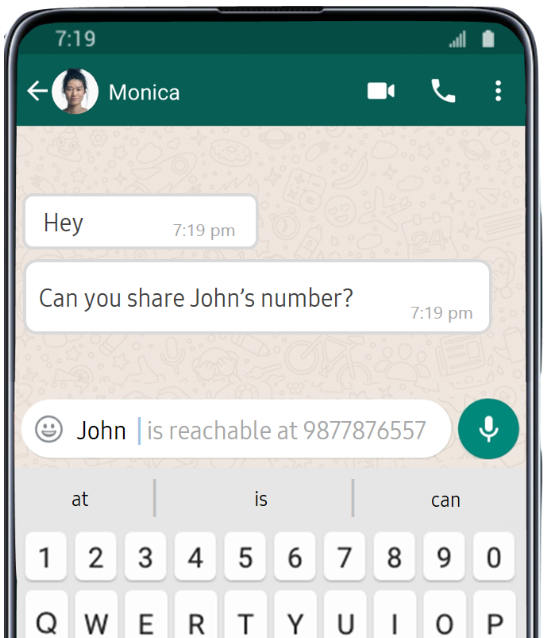}
		\label{fig:smartComposeSimple_b}
	\end{minipage}
	\caption{\textbf{Trigger-driven generation:} {\normalfont 
	User input is used as trigger to generate inline phrase completion.}}
	\label{fig:smartComposeSimple}
\end{figure}

Our motivating use cases involve resource-constrained edge devices.
When a smartphone/smartwatch user receives a message from a correspondent, he/she may find it helpful if a smart reply or action suggestion can be provided by a background service running on the phone. The smart suggestion may also need to leverage the data on the user's device. Due to latency expectations and obvious data privacy concerns, this processing must happen on-device. Furthermore, for the same data sample, different smart reply may be appropriate based on the intent of the incoming message, like replying with a message (Fig.~\ref{fig:teaser}(a)) or updating an existing schedule (Fig.~\ref{fig:teaser}(b)). On the other hand, during an active conversation, the user may appreciate if response suggestions can be easily personalized based on user-entered prefix token(s) (Fig.~\ref{fig:smartComposeSimple}).

To summarize, there is a need for data-to-text solutions that can (a) focus on a subset of a given data unit, (b) generate syntax-compliant text sequence(s), (c) emit alternatives based on the underlying intent, and (d) be flexible to adapt to user-provided triggers. In addition, such a solution should be lightweight and have low latency to be feasible for inference on edge devices. This motivates us to investigate the potential of a conditional attention-copy mechanism that not only allows us to identify the portion of data to be ``copied'' and accurately preserve unseen PIIs but also to predict adaptive and contextually relevant responses.

To this end, we propose TrICy, with an encoder-decoder foundation consisting of two encoders -- one focusing on the data dictionary and the other on the conditional drivers. Our primary contributions are:\looseness=-1

\begin{itemize}
    \item We introduce a novel framework, TrICy, that generates syntactically correct sequences, faithfully describing the relevant content in the input record by incorporating (a) an attention-copy mechanism, (b) message intent information, and (c) optional triggers.\looseness=-1
    
    \item We devise an efficient algorithm to determine the optimal ratio of introducing triggers in training data split to maximize the expected inference metrics.\looseness=-1
    
    \item Qualitative analysis of our proposal and comparison with the state-of-art (SOTA) shows that even with no conditional drivers, our method achieves SOTA-competitive performance with fewer parameters. Furthermore, to the best of our knowledge, TrICy is the first controlled data-to-text generation leveraging message intent and user-provided triggers, thereby delivering the new SOTA performance.\looseness=-1
    
\end{itemize}

\section{Related Work}\label{sec:relatedWork}

The concept of generating semantically coherent natural language descriptions from non-linguistic data has been explored since NLG became of interest among researchers worldwide. Owing to its versatile applicability, data-to-text generation has been researched and deployed in various domains including statistical data summarization \cite{iordanskaja1992generation}, health care \cite{gatt2009data}, stock market \cite{kukich1983design} and many more \cite{mei2015talk, leppanen2017data, van2017pass}. The early stages of research in this field involve simple rule-based approaches \cite{gatt2009data, gatt2009simplenlg, matheson2010moving}. Although these are efficient for simple tasks, they fail to work in complex scenarios \cite{mishra2020template}. A major drawback is that they require linguistic expert skills to formulate rules to achieve better performances.  \looseness=-1

\hfill

\paragraph{Sequence-to-Sequence paradigm}\label{sec:seq2seq}
The Seq2Seq paradigm is essentially based on the encoder-decoder framework. The emergence of Seq2Seq networks and methods led to a significant change in the quality of data-to-text generation and overcame shortcomings found in rule-based systems. 
Mei et al. \cite{mei2015talk} tackle the shortcomings of template-based approaches by introducing an LSTM-based encoder-aligner-decoder architecture, outperforming previous results by 58.8\% in the BLEU score on the Weather-Gov dataset \cite{liang2009learning}. Attention mechanisms \cite{bahdanau2014neural} were introduced in the following years, gaining immense popularity in neural network-based research.\looseness=-1

Seq2Seq models have also demonstrated their effectiveness on response generation task. To improve user experience with chatbots where typical response generation may lead to trivial conversations, topic-aware sequence-to-sequence architecture was introduced by Xing et al. \cite{xing2017topic}.
More informative and interesting responses are generated using a GRU based Seq2Seq architecture incorporating joint attention. Dziri et al. \cite{dziri2018augmenting} introduced a hierarchical joint attention neural model that considers both context and topical attention to produce diverse and contextual responses compared to the previous baselines. Although neural networks achieve remarkable results in this field of NLG, a significant setback in these approaches is the presence of many ``Unknown'' tokens in the final generated outputs, mainly due to the presence of Personally Identifiable Information (PII), which may be required to be replicated in the outputs as well.\looseness=-1

\hfill

\paragraph{Copying Mechanism in Seq2Seq learning}\label{sec:copymechanism}
CopyNet \cite{gu-etal-2016-incorporating} was introduced to overcome the limitations of simple neural networks for language generation. Following this advancement, further attempts to improve and fine-tune the copying mechanism for specific tasks have been widely explored.
Song et al. \cite{song2018structure} describes a method to incorporate and copy source syntactic structure to improve abstractive summarization. CopyRE \cite{zeng2018extracting} jointly extracts entities and relations from sentences using the proposed copy mechanism. Zeng et al.\cite{zeng2020copymtl} proposed CopyMTL, an advancement to CopyRE to enable multi-token entity prediction. Among recent works, CopyNext \cite{singh2020copynext} demonstrates a pipeline to copy continuous spans of text and shows its effectiveness for the task of nested named entity recognition. Bai et al. \cite{bai2021learning} proposed generating goal-oriented responses using a Seq2Seq pipeline incorporating a knowledge discerning module along with copy mechanism. The progress in the research of the copy mechanism further allowed a breakthrough to achieve SOTA performances in D2T generation.\looseness=-1

\hfill

\paragraph{Data-to-text Generation}\label{sec:d2tgeneratrion}
Lin et al. \cite{lin-etal-2020-data} introduced a pipeline to imitate the expected style and preserve content fidelity using a joint-attention with copy mechanism. This describes that using exemplar sentences as an input, the multi-encoder model learns to imitate the structure as they act as a soft template and demonstrate that their proposed architecture successfully beats previous attempts in data-to-text generation tasks. 
Similarly, Mishra et al. \cite{mishra2020template} uses automatically prepared POS-based templates for language generation from given keywords.
Yang et al. \cite{yang2021text} proposes a reinforcement learning based training strategy for data-to-text. However, it ignores the order of records for dynamic planning.

\hfill

Large Pre-trained Language Models (PLMs) have gained traction in D2T generation tasks \cite{lewis-etal-2020-bart, zhuang-etal-2021-robustly, castro-ferreira-etal-2020-2020}. Harkous et al. \cite{harkous-etal-2020-text} present a two-stage generation-reranking approach by utilizing fine-tuned GPT-2 model. However, to effectively utilize the PLMs for downstream tasks, these models are fine-tuned using task-specific objective functions \cite{zhou-srikumar-2022-closer, lewis-etal-2020-bart}. Recent generative language models, such as GPT-3 \cite{brown2020language}, ChatGPT \cite{chatgpt2022}, and Llama \cite{touvron2023llama1}, have been designed to learn and generate natural language text from multimodal inputs. These models are based on the transformer architecture that enables them to ingest large amounts of training data and perform well in zero-shot, fine-tuned settings. 
As we are further interested in enabling guided D2T, we take inspiration from prompt-based learning. This involves prepending task instructions to the input string and generating the outputs with its language modeling ability \cite{reif-etal-2022-recipe, li-liang-2021-prefix, carlsson-etal-2022-fine}. \looseness=-1

Although there is extensive research in the field of data-to-text generation, starting from simple rule-based approaches to heavy neural networks achieving SOTA, there has been little work done that focus on edge device deployment where balancing the trade-off between model size and performance is essential. Moreover, to our knowledge, generating contextual text sequences based on intent and user-provided trigger has not been explored. To this end, we investigate contextual sequence generation with TrICy and benchmark its performance on two public data-to-text datasets: E2E NLG \cite{novikova-etal-2017-e2e}, WebNLG \cite{gardent-etal-2017-creating}, and a Custom dataset, and demonstrate that our method outperforms competitive baselines.\looseness=-1

\section{Model Description}\label{sec:modelDescription}

We formulate our data-to-text generation task as:
\begin{equation}
	\mathcal{T} :
	\left\langle \mathcal{K}, \mathcal{I}, \overrightarrow{\mathcal{D}_c} \right\rangle
	\rightarrow
	\overrightarrow{\mathcal{S}_{g,c}}
\end{equation}
where in each input sample, $\mathcal{K}$ is a set of words, $\left\lbrace K_1, K_2, \cdots, K_{\left|\mathcal{K}\right|} \right\rbrace$, one of which can be an optional trigger input, $\mathcal{I}$ stands for intent class set, $\left\lbrace I_1, I_2, \cdots, I_{\left|\mathcal{I}\right|} \right\rbrace$, $\overrightarrow{\mathcal{D}_c}$ represents the set of ``context dictionary'' vectors, $\left\langle D_1, D_2, \cdots, D_{\left|F\right|} \right\rangle $, where each element $D_i$ denotes the input token sequence for field $f_i$ in an ordered field set $F$, and $\overrightarrow{\mathcal{S}_{g,c}}$ is the set of all text sequences, comprising of descriptions that expresses relations in $\overrightarrow{\mathcal{D}_c}$. Here, the subscripts $g$ and $c$ indicate the association of $\overrightarrow{\mathcal{S}_{g,c}}$ and $\overrightarrow{\mathcal{D}_c}$ with ``generate'' and ``copy'' that are discussed in Section \ref{sec:decodingWithCopyAndGenerate}.

\begin{figure}[tbp]
	\centering
	\includegraphics[width=\linewidth]{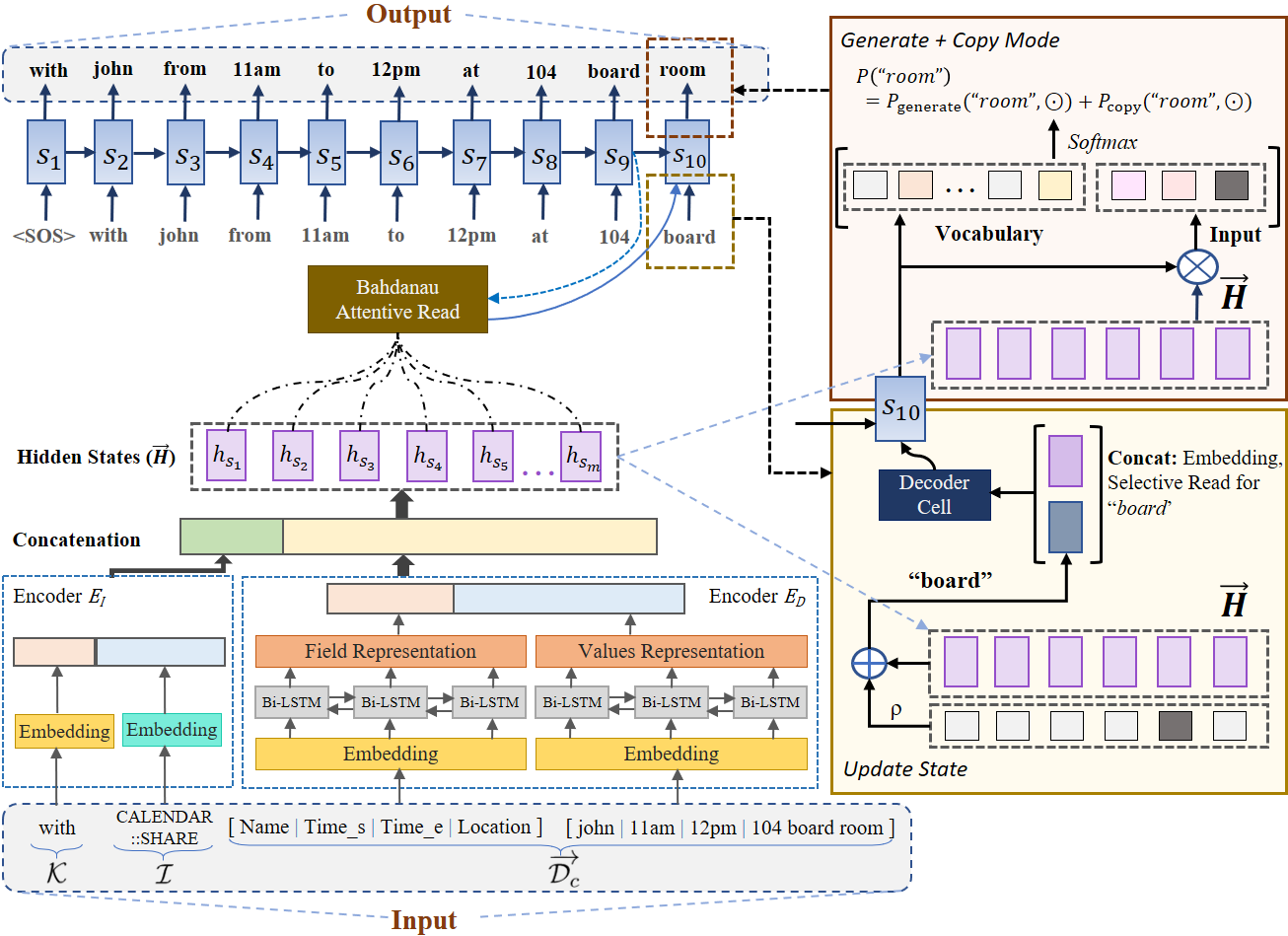}
	\caption{Architecture of the proposed TrICy Model}
	\label{fig:architecture}
\end{figure}

The TrICy architecture (Fig. \ref{fig:architecture}) consists of two encoders and one decoder. Encoder $E_I$ incorporates the intent information along with an optional trigger input, while encoder $E_D$ extracts the representation of input data record, namely fields, $f$, and their corresponding values, $v$.
An example of $f$ and $v$ in a meeting schedule data sample may include \big[~\textit{\texttt{Name}, ~\texttt{Time\_start}, ~\texttt{Time\_end}, ~\texttt{Location}}~\big] and \big[~\textit{``John'', ``11 am'', ``12 pm'', ``104 board room''}~\big] respectively.
The decoder conditions on the concatenated representation using distributed attention and predicts the target sequence using copy mechanism.\looseness=-1

\subsection{Encoders}\label{sec:encoders}

Each encoder reads different input data vectors to produce sequences of hidden states, i.e. hidden states from encoder $E_I$, $h_{E_I}$, and hidden states from encoder $E_D$, $h_{E_D}$.

The embedding layers in $E_D$ embed sequences of $N$ field tokens, $X_f = \left< x_{f_1}, x_{f_2}, \cdots, x_{f_N} \right>$, and value tokens, $X_v = \left< x_{v_1}, x_{v_2}, \cdots, x_{v_N} \right>$, by using contextual word embeddings. A bidirectional LSTM further encodes the input embeddings to learn input data representation. These hidden representations from $E_D$ are then concatenated with output from $E_I$ to get the final hidden memory state $\overrightarrow{H}$.\looseness=-1

\begin{equation}
	\begin{aligned}
		\overrightarrow{H} & = \left\langle h_{s_1}, h_{s_2}, \cdots, h_{s_m} \right\rangle \\
		                   & = \text{concat}\Big(\overrightarrow{h_{E_I}}, \overrightarrow{h_{E_D}}\Big)
	\end{aligned}
\end{equation}
where,
\begin{equation}
	\begin{aligned}
		\overrightarrow{h_{E_I}} & = \text{concat}\Big(	\overrightarrow{x_K}, \overrightarrow{x_I}\Big) \\
		\overrightarrow{h_{E_D}} & = \text{concat}
		    \left(\!\begin{aligned}
                &\text{BiLSTM}_1 \Big(\overrightarrow{x_{f_1}}, \overrightarrow{x_{f_2}}, \cdots, \overrightarrow{x_{f_N}} \Big),\\
                &\text{BiLSTM}_2 \Big(\overrightarrow{x_{v_1}}, \overrightarrow{x_{v_2}}, \cdots,\overrightarrow{x_{v_N}} \Big)
            \end{aligned}\right)
	\end{aligned}
\end{equation}
where, $\overrightarrow{x_{f_i}}$ is the word embedding for the field token, $x_{f_i}$, and $\overrightarrow{x_I}$ is the intent embedding vector.

\subsection{Bahdanau Attentive Read}\label{sec:bahdanauAttentiveRead}

The attention mechanism was first introduced in Seq2Seq to utilize the most relevant parts of the input sequence \cite{bahdanau2014neural, luong-etal-2015-effective}. Based on our ablation study (Section \ref{sec:ablation}), we utilize Bahdanau et al. \cite{bahdanau2014neural} in our final architecture, where we compute the context vector as a weighted sum of the encoder hidden states. Thus, we use a context vector, $c_t$, to encode the entire source words that the system attends to for generating the next word as defined below:\looseness=-1
\begin{equation}
	\begin{aligned}
		c_t &= \sum_{s=1}^{T} \alpha_{ts}h_s \\
		\quad \alpha_{ts} &= \frac{\exp\left(\mathcal{V}'\tanh\left(W_1s_{t-1} + W_2h_s\right)\right)}
		      {\sum_{s'}{\exp\left(\mathcal{V}'\tanh\left(W_1s_{t-1} + W_2h_{s'}\right)\right)}}
	\end{aligned}
\end{equation}
where, $h_s$ represents hidden states of the encoder, $s_{t-1}$ is decoder's previous time step state, and $W_1$, $W_2$ are weight matrices learnt during training via backpropagation.

\subsection{Decoding with Generate and Copy}\label{sec:decodingWithCopyAndGenerate}

As illustrated in Fig. \ref{fig:architecture}, TrICy is based on encoder-decoder framework and uses a copy mechanism \cite{gu-etal-2016-incorporating}. The inputs are transformed by Encoders (Section \ref{sec:encoders}) into representation $\overrightarrow{H}$ that is then read by the decoder to generate the output sequence. TrICy predicts output words based on the probability distribution of the generate mode and the copy mode, where the latter copies tokens from input sources (fields and values).

\paragraph{Decoder state}\label{sec:decoderState}

The predicted word at time $t-1$ is used to update the state at $t$. Calculating the decoder hidden state at each time step includes two attention mechanisms over the encoded source: attentive read and selective read. The attentive read is the Bahdanau et al. \cite{bahdanau2014neural} attention mechanism (as discussed in Section \ref{sec:bahdanauAttentiveRead}).

In selective read, we use weighted sum over the encoded source states as well as the
location-specific hidden state, $\overrightarrow{H}$, of the predicted word in previous time step. Thus, we use the copy probabilities assigned to each source token from the previous decoder step, normalize them to sum to one, and then use those to calculate a different weighted sum of the encoded states. Hence, the decoded state, $s_t$, is a function of the attentive read, selective read, and the predicted output token from the previous time step.\looseness=-1

\paragraph{Prediction}\label{sec:decoding}

The decoder utilizes $\overrightarrow{H}$ to predict the output sequence. We consider a vocabulary $\mathcal{V} = \left\lbrace x_1, x_2, \cdots, x_{\left|\mathcal{V}\right|} \right\rbrace$. Furthermore, we maintain unique words in the input source, $\mathcal{X}$, which may contain words, not in $\mathcal{V}$. Then, the extended vocabulary for source $\left(X_f \cup X_v\right)$ is $\left(\mathcal{V} \cup \mathcal{X}\right)$. In particular, the probability of a token $y_t$ from the extended vocab at the $t^{\text{th}}$ decoding step is given as:
\begin{equation}
	{
		P \left(y_t \mid \odot \right) = P_{\text{generate}} \left( y_t \mid \odot \right) + P_{\text{copy}} \left( y_t \mid \odot \right)
	}
\end{equation}
where, $\odot$ represents state of the decoder, which includes $s_{t-1}$, $y_{t-1}$, $c_t$, and $\overrightarrow{H}$. Let $\psi_{\text{generate}}$, $\psi_{\text{copy}}$ be score functions: $\psi_{\text{generate}}$ being the same as in the generic Bahdanau et al. \cite{bahdanau2014neural} RNN encoder-decoder, and $\psi_{\text{copy}}$ using a $\tanh$ transformation, as in Gu et al. \cite{gu-etal-2016-incorporating}.
Then, the final probability is:
\begin{equation}
	\begin{aligned}
		P \left(y_t \mid \odot \right) =
		\begin{cases}
			\frac{1}{Z} e^{\psi_{\text{generate}} \left( y_t \right)}                                & \text{if } y_t \in \mathcal{V}                         \\
			\frac{1}{Z} \sum\limits_{j: x_j = y_t} e^{\psi_{\text{copy}} \left( x_j \right)} & \text{if } y_t \in \mathcal{X} - \mathcal{V}
		\end{cases}
	\end{aligned}
    \label{eq:finalProbability}
\end{equation}
where $Z$ is a normalizing factor, as given by:
\begin{equation}
    Z = \sum_{v \in \mathcal{V}} e^{\psi_{\text{generate}}\left(v\right)} + \sum_{x \in \mathcal{X}} e^{\psi_{\text{copy}}\left(x\right)}
\end{equation}

The two decoding modes compete through a $\softmax$ ($\sigma$) function due to the normalization term. Further, we add the probabilities of all $x_j$ equal to $y_t$ in Equation \eqref{eq:finalProbability} considering that there may be multiple input words for decoding $y_t$. It may be noted that $P_{\text{copy}} \left( y_t \mid \odot \right) = 0$ if $y_t$ is not present in the input words, whereas $P_{\text{generate}} \left( y_t \mid \odot \right) = 0$ when $y_t$ only appears in the input.\looseness=-1

\subsection{Trigger input, \texorpdfstring{$\mathcal{K}$}{K}}\label{sec:triggerInput}
To further control D2T generation, we introduce an optional input, namely, trigger. Trigger is intended to be an input feature during training, and hence, comprises of vocabulary words. Moreover, it leads the output sequence as in Fig. \ref{fig:smartComposeSimple}. 
This is in contrast to prompting that typically alters the behavior of PLMs \cite{carlsson-etal-2022-fine}, seldom plays a role in their training \cite{https://doi.org/10.48550/arxiv.2107.13586}, and needs to be engineered either manually or automatically \cite{li-liang-2021-prefix}.\looseness=-1

By adding a text token as an additional input for a portion of the training set, we train the model to get directed predictions. We define the ratio of training samples that are augmented with trigger tokens as ${}_tr_\mathcal{K}$, and that of test or evaluation samples as ${}_er_\mathcal{K}$. In our implementation, we use the first word of $\overrightarrow{\mathcal{S}_{g,c}}$ as the trigger in the training set. For the samples that are left unaugmented, we use a special token, ``\texttt{<SOS>}'' as a start-of-sentence placeholder. We present an algorithm to compute the optimal setting for the usage of triggers (${}_tr_\mathcal{K}^*$) in Section \ref{sec:triggerRatioExperiments}.
\looseness=-1

\section{Experimental Setup}\label{sec:experimentalSetup}

\subsection{Dataset}\label{sec:dataset}

We use two public datasets and one Custom dataset for evaluation, as summarized in Table \ref{tab:datasetStatistics}:

	\paragraph{E2E NLG \cite{novikova-etal-2017-e2e}} The dataset contains pairs of meaning representations (MR) and utterances in the restaurant domain, where an MR can include a subset of 8 fields.

\begin{table}[tbp]
	\caption{Dataset Statistics}
	\centering
	\resizebox{\linewidth}{!}{%
		\begin{tabular}{c r r r c c r}
			\toprule
			\multirow{2}{*}{\textbf{Dataset}} &    \multicolumn{3}{c}{\textbf{Number of instances}}     & \multicolumn{1}{c}{\textbf{No. of}}     & \multicolumn{1}{c}{\textbf{No. of}}  & \multicolumn{1}{c}{\textbf{Vocabulary}} \\ \cline{2-4}
			                                  & \textbf{Training} & \textbf{Validation} & \textbf{Test} & \multicolumn{1}{c}{\textbf{attributes}} & \multicolumn{1}{c}{\textbf{intents}} & \multicolumn{1}{c}{\textbf{size}}       \\ \midrule
			               E2E                & 42061             & 4672                & 4693          & 8                                       & {\color{gray}N/A}                    & 2620                                    \\
			               {WebNLG}             & 13211             & 1667                & 1779          & 372\footnotemark\saveFN\AttributeCount\                                      & 16\footnotemark\saveFN\IntentCount\                    & 9395                                   \\
			            Custom             & 14470              & 2350                 & 2390           & 7                                       & 10                                    & 542                                     \\ \bottomrule
			            \multicolumn{7}{l}{{\small \useFN\AttributeCount\ Attributes seen in training set}}\\
			            \multicolumn{7}{l}{{\small \useFN\IntentCount\ Distinct \textit{``category''} attributes in training set}}\\
		\end{tabular}
	}
	\label{tab:datasetStatistics}
\end{table}

\begin{table}[bp]
	\caption{
	    \textbf{Distribution of Custom dataset:}
	    {\small 
	        $(I+D)$s are combination aliases later referred in Section \ref{sec:triggerRatioExperiments}.
	    }
	}
	\centering
	\resizebox{\linewidth}{!}{%
		\begin{tabular}{c c c c}
			\toprule
			$(I+D)$ alias & $\mathcal{I}$ & \textbf{Populated $f_i$s in $\overrightarrow{\mathcal{D}_c}$} & \textbf{\#Samples}\\ \midrule \midrule
			${(I+D)}_{1}$ & \multirow{3}{*}{CONTACT::ACT} & Contact, Email, Name  & 875 \\
            ${(I+D)}_{2}$ &                               & Contact, Name  & 975 \\
            ${(I+D)}_{3}$ &                               & Email, Name  & 750 \\
            \midrule
            ${(I+D)}_{4}$ & \multirow{3}{*}{CONTACT::SHARE} & Contact, Email, Name  & 875 \\
            ${(I+D)}_{5}$ &                                 & Contact, Name  & 975 \\
            ${(I+D)}_{6}$ &                                 & Email, Name  & 750 \\
            \midrule
            ${(I+D)}_{7}$ & LOCATION::SHARE & Location & 570\\
            \midrule
            ${(I+D)}_{8}$ & \multirow{6}{*}{CALENDAR::ACT} & Date, Location, Title, Name, ID, TimeStart, TimeEnd  & 2210 \\
            ${(I+D)}_{9}$  &                              & Date, Location, Title, ID, TimeStart, TimeEnd  & 680 \\
            ${(I+D)}_{10}$ &                              & Date, Title, ID, TimeStart, TimeEnd  & 470 \\
            ${(I+D)}_{11}$ &                              & Location, Title, Name, ID  & 495 \\
            ${(I+D)}_{12}$ &                              & Location, Name, ID  & 200 \\
            ${(I+D)}_{13}$ &                              & Name, ID, TimeStart  & 200 \\
            \midrule
            ${(I+D)}_{14}$ & \multirow{6}{*}{CALENDAR::SHARE} & Date, Location, Title, Name, ID, TimeStart, TimeEnd  & 2210 \\
            ${(I+D)}_{15}$ &                                 & Date, Location, Title, ID, TimeStart, TimeEnd  & 680 \\
            ${(I+D)}_{16}$ &                                 & Date, Title, ID, TimeStart, TimeEnd  & 470 \\
            ${(I+D)}_{17}$ &                                 & Location, Title, Name, ID  & 495 \\
            ${(I+D)}_{18}$ &                                 & Location, Name, ID  & 200 \\
            ${(I+D)}_{19}$ &                                 & Name, ID, TimeStart  & 200 \\
            \midrule
            ${(I+D)}_{20}$ & OCCASION::SHARE & Name, Occasion  & 2040 \\
            \midrule
            ... & ...                          & ...  & \multirow{2}{*}{\bigg\}2890} \\
            ... & ...                          & ...  &  \\
            \bottomrule
			
		\end{tabular}
	}
	\label{tab:heatmapDescription}
\end{table}

\definecolor{cell_none}{rgb}{0,0,0}
\definecolor{cell_bad}{rgb}{1,.95,.95}
\definecolor{cell_neutral}{rgb}{1,1,0.9}
\definecolor{cell_good}{rgb}{.95,1,.95}

\definecolor{dt_name}{rgb}           { 0,    0, .75}
\definecolor{dt_eatType}{rgb}        { 0,  .75,   0}
\definecolor{dt_food}{rgb}           { 0,  .75, .75}
\definecolor{dt_area}{rgb}           {.75,   0,   0}
\definecolor{dt_familyFriendly}{rgb} {.75,   0, .75}
\definecolor{dt_near}{rgb}           {.75,  .5,   0}

\begin{table*}[tb]
    \setlength{\arrayrulewidth}{1pt}
	\caption{\textbf{Generated text sequences and aggregated BLEU scores for E2E NLG dataset.}
	{\small \normalfont 
	Depicted triggers, $\mathcal{K}$, are representative of 4 unique classes (based on whether they belong to vocabulary, $\mathcal{V}$, and the set of gold references,  $\protect\overrightarrow{\mathcal{S}_{g,c}}$, corresponding to the dictionary record, $\in \protect\overrightarrow{\mathcal{D}_c}$).
	Model parameters (in millions) are denoted by $\Pi$. Values in parentheses denote BLEU scores averaged over all test samples, with 3 random triggers for each class of $\mathcal{K}$. Input for intent $\mathcal{I}$ is set to a constant dummy value, $I_{\text{0}}$, as it is absent in the dataset.
	(Effect of $\mathcal{I}$ can be observed in Tables \ref{tab:sotaEvalWebNLG} and \ref{tab:qualitativeResults} with WebNLG and Custom dataset.)
	}}
	\centering
	\resizebox{\linewidth}{!}{%
		\begin{tabular}{P{0.1\linewidth} | p{0.18\linewidth} | p{0.12\linewidth} | p{0.12\linewidth} | p{0.15\linewidth} | p{0.18\linewidth} | p{0.18\linewidth} | p{0.18\linewidth} }
			\toprule
			
			\raggedleft{$\protect\overrightarrow{\mathcal{D}_c}$} & \multicolumn{5}{l}{\textit{\{{\color{dt_name} name: ``Green Man''}, {\color{dt_eatType} eatType: ``pub''}, {\color{dt_food} food: ``Italian''}, {\color{dt_area} area: ``riverside''}, {\color{dt_familyFriendly} familyFriendly: ``yes''}, {\color{dt_near} near: ``Express by Holiday Inn''}\}}} 
			\\ \midrule
			
			\raggedleft{$\protect\overrightarrow{\mathcal{S}_{g,c}}$} & \multicolumn{6}{l}{\textit{%
			``{\color{dt_name} Green Man} is a {\color{dt_eatType} pub} located by the {\color{dt_area} river} near the {\color{dt_near} Express by Holiday Inn}. It serves {\color{dt_food} Italian} {\color{dt_food} food} and is {\color{dt_familyFriendly} family friendly}.''%
			} + 34 alternatives
			\cite{novikova-etal-2017-e2e}}
			\\ \midrule\midrule
			
			\raggedleft{Model}
			& 
			Llama 2 \cite{touvron2023llama} 7B (Fine-tuned\footnotemark\saveFN\LlamaETwoE\ )
			&
			Elder et al. \cite{elder-etal-2018-e2e}
			& 
			Zhang et al. \cite{zhang2018attention}
			& 
			Shen et al. \cite{shen-etal-2019-pragmatically}
			&
			TrICy (${}_tr_\mathcal{K} = 0$), $M_7$
			& 
			TrICy (${}_tr_\mathcal{K} = 1$), $M^{'}_7$
			& 
			TrICy (${}_tr_\mathcal{K}$ = ${}_tr_\mathcal{K}^* = 0.65$), $M^{*}_7$
			\\ \midrule 
			
			{\backslashbox{$\mathcal{K}$}{$\Pi$}}
			& 
			\multicolumn{1}{c|}{
			    $7000.0$ M
			}
			& 
			\multicolumn{1}{c|}{
			    -
			}
			& 
			\multicolumn{1}{c|}{
			    $87.0$ M
			}
			&
			\multicolumn{1}{c|}{
			    $5.1$ M
			}
			&
			\multicolumn{3}{c}{
			    $\mathbf{4.7}$ \textbf{M}
			}
			\\ \midrule 
			
			{``\texttt{<SOS>}'' \newline Class $\mathcal{C}_1$: No trigger}
			&
			\cellcolor{cell_bad} (38.47) \textit{The {\color{dt_name} Green Man} {\color{dt_eatType} pub} is located {\color{dt_near} near} {\color{dt_near} Express by Holiday Inn} and is {\color{dt_familyFriendly} family friendly}.}
			&
			\cellcolor{cell_neutral} (65.93) \textit{{\color{dt_name} Green Man} is a Children friendly {\color{dt_food} Italian} {\color{dt_eatType} pub} {\color{dt_near} near} {\color{dt_near} Express by Holiday Inn} in {\color{dt_area} riverside}. }
			&
			\cellcolor{cell_neutral} (65.45) \textit{{\color{dt_name} Green Man} is a {\color{dt_familyFriendly} family friendly} {\color{dt_food} Italian} {\color{dt_eatType} pub} in the {\color{dt_area} riverside} {\color{dt_area} area} {\color{dt_near} near} {\color{dt_near} Express by Holiday Inn}. }
			&
			\cellcolor{cell_good} (\textbf{68.60}) \textit{{\color{dt_name} Green Man} is a {\color{dt_familyFriendly} family friendly} {\color{dt_eatType} pub} that also serves {\color{dt_food} Italian} {\color{dt_food} food} and is near the {\color{dt_area} riverside} and the {\color{dt_near} Express by Holiday Inn}.}
			& 
			\cellcolor{cell_good} (66.43) \textit{{\color{dt_name} Green Man} is a {\color{dt_familyFriendly} family friendly} {\color{dt_food} Italian} {\color{dt_eatType} pub}. It is along the {\color{dt_area} riverside} {\color{dt_near} near} {\color{dt_near} Express by Holiday Inn}. }
			& 
			\cellcolor{cell_bad} (60.26) \textit{{\color{dt_name} Green Man} is a {\color{dt_familyFriendly} family friendly} {\color{dt_eatType} pub} that specializes in {\color{dt_food} Italian} {\color{dt_food} food}. It {\color{dt_near} near} {\color{dt_near} Express by Holiday Inn}. }
			&
			\cellcolor{cell_good} (67.82) \textit{{\color{dt_name} Green Man} is a {\color{dt_familyFriendly} family friendly} {\color{dt_eatType} pub} that specializes in {\color{dt_food} Italian} {\color{dt_food} food}. It is along the {\color{dt_area} riverside} {\color{dt_near} near} {\color{dt_near} Express by Holiday Inn}. }
			\\ \midrule
			
			{``\underline{ghgsdp..}'' \newline (garbage token) \newline Class $\mathcal{C}_2$: \newline $\notin \mathcal{V}$}
			&
			\cellcolor{cell_bad} (-) \textit{\underline{Ghgsdplgkjbq}} \newline {\color{gray} (No further response generated before ``\texttt{<END>}''. Tokens like ``qwerty'' are in Llama 2 vocabulary, and hence, not used for $\mathcal{K}$.)}
			& 
			\color{gray}\centering{N/A} 
			& 
			\color{gray}\centering{N/A} 
			&
			\color{gray}\centering{N/A} 
			& 
			\cellcolor{cell_neutral} (65.93) \textit{{\color{dt_name} Green Man} is a {\color{dt_familyFriendly} family friendly} {\color{dt_food} Italian} {\color{dt_eatType} pub}. It is along the {\color{dt_area} riverside} {\color{dt_near} near} {\color{dt_near} Express by Holiday Inn}. }
			& 
			\cellcolor{cell_bad} (59.13) \textit{{\color{dt_name} Green Man} is a {\color{dt_familyFriendly} family friendly} {\color{dt_eatType} pub} that specializes in {\color{dt_food} Italian} {\color{dt_food} food}. It {\color{dt_near} near} {\color{dt_near} Express by Holiday Inn}. }
			& 
			\cellcolor{cell_good} (\textbf{67.80}) \textit{{\color{dt_name} Green Man} is a {\color{dt_familyFriendly} family friendly} {\color{dt_eatType} pub} that specializes in {\color{dt_food} Italian} {\color{dt_food} food}. It is along the {\color{dt_area} riverside} {\color{dt_near} near} {\color{dt_near} Express by Holiday Inn}. }
			\\ \midrule
			
			{``\underline{with}'' \newline Class $\mathcal{C}_3$: \newline $\in \mathcal{V}$ \newline $\land$ \newline $\notin \protect\overrightarrow{\mathcal{S}_{g,c}}$}
			&
			\cellcolor{cell_bad} (42.58) \textit{\underline{With} a kids-friendly atmosphere, the {\color{dt_name} Green Man} {\color{dt_eatType} pub} serves {\color{dt_food} Italian} {\color{dt_food} food} and is located in {\color{dt_area} riverside} {\color{dt_near} near} {\color{dt_near} Express by Holiday Inn}. }
			& 
			\color{gray}\centering{N/A} 
			& 
			\color{gray}\centering{N/A} 
			&
			\color{gray}\centering{N/A} 
			&
			\cellcolor{cell_neutral} (65.95) \textit{{\color{dt_name} Green Man} is a {\color{dt_familyFriendly} family friendly} {\color{dt_eatType} pub} in {\color{dt_area} riverside} that serves {\color{dt_food} Italian} {\color{dt_food} food}. }
			& 
			\cellcolor{cell_good} (67.90) \textit{\underline{With} a nice {\color{dt_familyFriendly} family friendly} atmosphere, {\color{dt_name} Green Man} is a {\color{dt_familyFriendly} family friendly} {\color{dt_eatType} pub} located {\color{dt_near} near} {\color{dt_near} Express by Holiday Inn} in the {\color{dt_area} riverside} {\color{dt_area} area}. }
			& 
			\cellcolor{cell_good} (\textbf{69.21}) \textit{\underline{With} a nice {\color{dt_familyFriendly} family friendly} atmosphere, {\color{dt_name} Green Man} is a {\color{dt_food} Italian} {\color{dt_eatType} pub} located {\color{dt_near} near} {\color{dt_near} Express by Holiday Inn} in the {\color{dt_area} riverside} {\color{dt_area} area}. }
			\\ \midrule
			
			{``\underline{near}'' \newline Class $\mathcal{C}_4$: \newline $\in \mathcal{V}$ \newline $\land$ \newline $\in \protect\overrightarrow{\mathcal{S}_{g,c}}$}
			&
			\cellcolor{cell_bad} (43.04) \textit{\underline{{\color{dt_near} Near}} {\color{dt_near} Express by Holiday Inn} is a {\color{dt_familyFriendly} family friendly} {\color{dt_eatType} pub} called {\color{dt_name} Green Man}.}
			& 
			\color{gray}\centering{N/A} 
			&
			\color{gray}\centering{N/A} 
			&
			\color{gray}\centering{N/A} 
			&
			\cellcolor{cell_neutral} (65.95) \textit{{\color{dt_name} Green Man} is a {\color{dt_familyFriendly} family friendly} {\color{dt_eatType} pub} in {\color{dt_area} riverside} that serves {\color{dt_food} Italian} {\color{dt_food} food}. }
			& 
			\cellcolor{cell_good} (68.08) \textit{\underline{{\color{dt_near} Near}} {\color{dt_near} Express by Holiday Inn} is {\color{dt_name} Green Man} which is a {\color{dt_familyFriendly} family friendly} {\color{dt_eatType} pub} serving {\color{dt_food} Italian} {\color{dt_food} food}. It is along the {\color{dt_area} riverside}. }
			&
			\cellcolor{cell_good} (\textbf{69.29}) \textit{\underline{{\color{dt_near} Near}} {\color{dt_near} Express by Holiday Inn} is a {\color{dt_name} Green Man} {\color{dt_familyFriendly} family friendly} {\color{dt_eatType} pub} serving {\color{dt_food} Italian} {\color{dt_food} food}. It is along the {\color{dt_area} riverside} {\color{dt_area} area}. }
			\\ \bottomrule
			
			\multicolumn{8}{l}{\useFN\LlamaETwoE\ Details about fine-tuning configuration and prompt design are presented in Appendix \ref{sec:LlamaFineTuningParameters}.}
			
		\end{tabular}
	}
	\label{tab:sotaQualitativeE2E}
\end{table*}

    \paragraph{WebNLG \cite{gardent-etal-2017-creating}} This consists of triplet sets and corresponding natural language descriptions. We validate our model on RDF-to-text generation and use version 3.0 for our work. The training set contains up to 7 triplets each, corresponding to 16 DBpedia topics and one or more reference texts per set. The test set is split into three parts: \textit{seen}, \textit{unseen entities}, and \textit{unseen}. We use the \textit{``category''} attribute as a proxy for intent input and evaluate our model on \textit{seen} and \textit{unseen} categories.\looseness=-1
	
	\paragraph{Custom dataset} With the motivation to build an on-device contextual reply  (Fig. \ref{fig:teaser} and \ref{fig:smartComposeSimple}), we curate a dataset where each data sample consists of (a) one intent class, $\in \mathcal{I}$, (ex: \texttt{CONTACT::SHARE}), (b) context information, $\in \overrightarrow{\mathcal{D}_c}$, (ex: \textit{``Name[Lucas], Email[lucas@example.org]''}), and (c) target response sequence, $\in \overrightarrow{\mathcal{S}_{g,c}}$, (ex: \textit{``You can mail Lucas at lucas@example.org''}). Table \ref{tab:heatmapDescription} illustrates the  distribution of this dataset and a few samples from each $(I+D)$ can be found in Appendix \ref{sec:CustomDatasetSamples}.\looseness=-1
	
	\paragraph*{\textbf{Curation process}} For the custom data curation phase, context dictionary field-value pairs are generated for all possible $(\mathcal{I} + \mathcal{D})$ combinations. Target sequences for \texttt{*::SHARE} intents are then aggregated from $38$ volunteers from the same subcontinent, who have consented to contribute. Each user is presented $\sim65$ random pairs of intents and dictionary records, for each of which they prepare a natural language sentence that captures all relations in $\overrightarrow{\mathcal{D}_c}$ (or an almost-complete subset of it). 133 sentences are discarded because of one or more reasons: (a) covers $<80\%$ of dictionary values, (b) has offensive word(s) from a predetermined list, (c) contains named entities $\notin \overrightarrow{\mathcal{D}_c}$. These sentences are then used to generate four augmented references by selecting a new dictionary record with the same field subset $\in F$ but different corresponding values. Corresponding to each of these inputs, wherever applicable, \texttt{*::ACT} sentences, dealing with markup sequences intended for downstream tasks such as rendering actionable UI (Fig. \ref{fig:teaser}b), are generated programmatically by populating placeholders with the dictionary record values.

\subsection{Implementation Details}\label{sec:implementationDetails}

We train all the models on an NVIDIA GeForce GTX 1080 Ti GPU with 12GB RAM using TensorFlow \cite{abadi2016tensorflow} framework. The embedding dimensions to encode field and intent are set to 16 each, and those for value and trigger are each set to 50. We use a subset of the 50-dim GloVe \cite{pennington-etal-2014-glove} embeddings corresponding to the training set vocabulary. The batch size is set to 64, and we apply a regular dropout of 0.2 to reduce model overfitting with the Custom dataset. We experiment with attention mechanisms using 256 units as depth and set the cell size of the decoder and CopyNet \cite{gu-etal-2016-incorporating} layer to 256 units. We use the Adam optimizer \cite{Kingma2015AdamAM} with a constant learning rate of 0.001 and set the maximum epochs to 18. The validation loss after each epoch is used to select the best model. We apply beam search decoding and set beam width as 3.\looseness=-1

\section{Evaluation Results}\label{sec:evaluationResults}

At the outset, we present a comparison of TrICy models with SOTA models in section \ref{sec:comparisonWithSOTA}. Then, we present an analysis of our architecture components through an ablation study in section \ref{sec:ablation}. We showcase how the optimal variant of TrICy, $M^{*}_7$, is arrived at in section \ref{sec:triggerRatioExperiments}. Thereafter, we present further analysis and insights in the subsequent sections.

\subsection{Comparison with SOTA}\label{sec:comparisonWithSOTA}

\subsubsection{On E2E NLG dataset}\label{sec:resultsOnE2E}

Table \ref{tab:sotaQualitativeE2E} presents a sample of text sequences generated by TrICy and related work \cite{touvron2023llama,elder-etal-2018-e2e, zhang2018attention, shen-etal-2019-pragmatically} when trained on E2E dataset. Among LLMs, we select Llama 2 as it the weights are open-source. Instead of exploring a variety of advanced prompt engineering methods, which have been shown to be less effective with structured data \cite{masson2023directgpt}, we select a simplistic prompt capturing the values in $\overrightarrow{\mathcal{D}_c}$, as the effectiveness of this approach for D2T tasks has been established by Ferreira et al. \cite{castro-ferreira-etal-2020-2020}. We fine-tune the Llama 2 7B model on the train split \cite{novikova-etal-2017-e2e}, on an NVIDIA RTX A6000 GPU with 48 GB RAM, with the configuration detailed in Appendix \ref{sec:LlamaFineTuningParameters}.

From the aggregated BLEU scores in Table \ref{tab:sotaQualitativeE2E}, we observe that when trigger is absent in test samples (i.e. class $\mathcal{C}_1$), our $M_7$ model (introduced in Table \ref{tab:ablationResults}) performs better than the former two approaches and comparable to the previously known SOTA\footnote{https://paperswithcode.com/sota/data-to-text-generation-on-e2e-nlg-challenge (accessed on 25-Sep-2023)} for this dataset \cite{shen-etal-2019-pragmatically}. We show that, as expected, training TrICy with triggers (using leading words from gold reference) supplied for all training samples ($M^{'}_7$) leads to a sharp drop in $\mathcal{C}_1$ as the model heavily relies on the presence of trigger in input which is absent in test set. However, our best TrICy model, $M^{*}_7$, trained with the optimal training ratio (from Fig. \ref{fig:lineGraphBleuE2E}) achieves a BLEU score of 67.82\% for $\mathcal{C}_1$, which is very close to that of Shen et al. \cite{shen-etal-2019-pragmatically} with fewer model parameters.\looseness=-1

In the case of $\mathcal{C}_2$, we attribute the minor score drop to the fact that OOV tokens are rarer than ``\texttt{<SOS>}'' tokens. Thus, for practical purposes, OOVs in triggers may be safely replaced with ``\texttt{<SOS>}''. With $\mathcal{C}_3$ and $\mathcal{C}_4$, $M^{'}_7$ outperforms $M_7$ due to the presence of triggers in training data for $M^{'}_7$. Furthermore, $M^{*}_7$ outperforms both $M_7$ and $M^{'}_7$ due to its optimal reliance on triggers and consequent ability to ignore triggers when they are inconsistent with the rest of the input values. Interestingly, we observe that fine-tuned Llama 2 performs worst in all settings. This indicates that while pre-trained generative models generate meaningful information, they may miss out on the primary goal of preserving content fidelity in D2T tasks. Finally, we observe that in the most typical case $\mathcal{C}_4$, $M^{*}_7$ produces the new SOTA value of 69.29\% with lowest memory footprint and without having used any additional training data.\looseness=-1

\subsubsection{On WebNLG dataset}\label{sec:resultsOnWebNLG}

We present a performance comparison on the WebNLG dataset in Table \ref{tab:sotaEvalWebNLG}. 
Since BLEU is precision-oriented and may be unable to capture certain nuances of generation quality, we also use METEOR for evaluation as it has a better correlation with human judgment. This is because METEOR is a weighted F-score and uses stemming and synonym matching to align words between candidate and reference \cite{hadla2015comparative}. For D2T models like BART \cite{lewis-etal-2020-bart}, we leverage TextBox 2.0 \cite{tang2022textbox} implementation due to its improvements. We also include LLMs like GPT-3, ChatGPT, and Llama 2, in our comparison. 
While the experiments with GPT-3 and ChatGPT are performed using foundational models in zero-shot setting, Llama~2 is fined-tuned on the dataset.
For ChatGPT, we refer to the configuration used by the Axelsson et al. \cite{axelsson-skantze-2023-using} as it demonstrates superior performance and observes less hallucinatory text continuation, compared to contemporary work like Yuan et al. \cite{yuan-faerber-2023-evaluating} that achieves a BLEU score of 11.08. Nevertheless, we observe similar drawbacks with LLMs for D2T task as seen in Section \ref{sec:resultsOnE2E}. TrICy achieves SOTA-competitive scores while having the least number of parameters.

\begin{table}[tbp]
	\caption{\small Performance on WebNLG dataset ($\Pi$: Model parameters)
	}
	\centering
	\resizebox{\linewidth}{!}{%
		\begin{tabular}{c | c | r | c | c | c | c}
			\toprule
			
			\multirow{2}{*}{\textbf{Model}}
			& \multirow{2}{*}{\textbf{Base}}
			& \multicolumn{1}{c|}{$\mathbf{\Pi}$ }
			& \multicolumn{2}{c|}{\textbf{BLEU (\%)}}
			& \multicolumn{2}{c}{\textbf{METEOR (\%)}}
			\\ \cline{4-7}
			
			&
			& \multicolumn{1}{c|}{\textbf{(million)}}
			& \textbf{Seen}
			& \textbf{Unseen}
			& \textbf{Seen}
			& \textbf{Unseen}
			\\\midrule\midrule
			
			GPT-3 (Zero-shot) \cite{brown2020language}
			& text-davinci-003
			& 175000
			& \multicolumn{2}{c|}{20.36\footnotemark}
			& \multicolumn{2}{c}{26.95\footnotemark[\value{footnote}]\saveFN\GPT\ }
			\\
			
			ChatGPT (Zero-shot) \cite{chatgpt2022}
			& gpt-3.5-turbo-0301
			& Unknown
			& \multicolumn{2}{c|}{42.40\footnotemark}
			& \multicolumn{2}{c}{40.90\footnotemark[\value{footnote}]\saveFN\ChatGPT\ }
			\\
			
			Llama 2 (Fine-tuned) \cite{touvron2023llama}
			& Llama-2-7b
			& 7000
			& 48.54\footnotemark
			& 42.61\footnotemark[\value{footnote}]
			& 41.33\footnotemark[\value{footnote}]
			& 39.82\footnotemark[\value{footnote}]\saveFN\LlamaWebNLG\
			\\
			
			Clive et al. \cite{https://doi.org/10.48550/arxiv.2110.08329}
			& T5-Large
			& $>$770
			& \textbf{67.32}
			& 56.41
			& -
			& -
			\\
			
			Wang et al. \cite{wang-etal-2021-stage}
			& T5-Large
			& $>$770
			& 66.07
			& 53.87
			& \textbf{46.00}
			& 42.00
			\\
			
			Kale and Rastogi \cite{kale-rastogi-2020-text}
			& T5-Base
			& 220
			& 64.70
			& 52.80
			& \textbf{46.00}
			& 41.00
			\\
			
			TextBox 2.0 \cite{tang2022textbox}
			& BART
			& $>$140
			& -
			& \textbf{67.33}
			& -
			& \textbf{47.78}
			\\
			
			Ribeiro et al. \cite{https://doi.org/10.48550/arxiv.2001.11003}
			& CGE-LW
			& 10.5
			& 63.69
			& -
			& 44.47
			& -
			\\
			
			TrICy (${}_tr_\mathcal{K} \newline = 0$)
			& None
			& \textbf{6.2}
			& 64.08
			& 52.35
			& 45.23
			& 40.82
			\\
			
			TrICy (${}_tr_\mathcal{K} \newline = {}_tr_\mathcal{K}^* \newline = 0.24$)
			& None
			& \textbf{6.2}
			& 64.73
			& 52.91
			& 45.53
			& 41.16
			\\ \bottomrule
			
			\multicolumn{7}{l}{
			\useFN\GPT\ \textsuperscript{,}\useFN\ChatGPT\ \textsuperscript{,}\useFN\LlamaWebNLG\ Prompt design and configuration are as in \cite{yuan-faerber-2023-evaluating}, \cite{axelsson-skantze-2023-using}, and Appendix \ref{sec:LlamaFineTuningParameters}, respectively.
			}
			
		\end{tabular}
	}
	\label{tab:sotaEvalWebNLG}
\end{table}

\hfill

\paragraph*{\textbf{A note on efficiency}}\label{sec:noteOnEfficiency}
TrICy has $<60$\% of parameters compared to the smallest competitor, Ribeiro et al. \cite{https://doi.org/10.48550/arxiv.2001.11003} that has a model size of 130 MB. We further apply post-training Float16 quantization\footnote{https://www.tensorflow.org/lite/performance/post\_training\_float16\_quant (accessed on 12-Feb-2023)} before deploying our model on-device.
Moreover, even though GPU is utilized for model training, the encoder-decoder architecture is light enough to be trainable on a Intel{\sffamily\textregistered}\hspace{0.2mm} Xeon{\sffamily\textregistered}\hspace{0.2mm} Silver 4210 CPU @ 2.20GHz without any dedicated GPU.
On Custom dataset, TrICy has an average on-device inference latency of 43~ms on a Samsung Galaxy S20 device (8 GB RAM, 2 GHz octa-core Exynos 990 processor).

\subsection{Ablation Study}\label{sec:ablation}

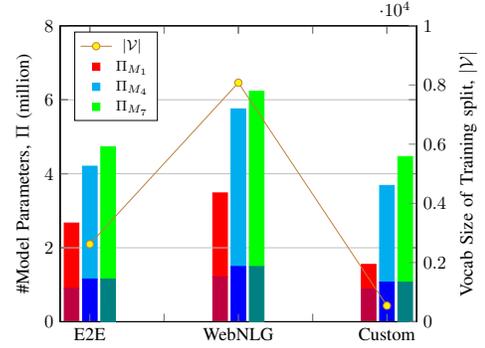
\begin{figure}[tb]
	\centering
	\resizebox{.72\linewidth}{!}{%
		\begin{tikzpicture}[
		    yscale=1,
		    /pgfplots/every axis/.style={ %
            symbolic x coords={
				{E2E},
				{WebNLG},
				{Custom}
			},
            bar width=8pt,
            ymin=0, ymax=8.000000,
          },
		]
			\begin{axis}[
				ybar stacked,
				bar shift=-10pt,
				xticklabels={},
				yticklabels={},
				]
				\addplot+[draw=purple,fill=purple] coordinates {
					(E2E, 0.904192) (WebNLG, 1.2090746) (Custom, 0.8836192)
				};
				\addplot+[draw=red,fill=red] coordinates {
					(E2E, 1.7635904) (WebNLG, 2.2730788) (Custom, 0.66616)
				};
				\label{plot_y_mp_m1}
			\end{axis}
			
			\begin{axis}[
				ybar stacked,
				bar shift=0pt,
				xtick=data,
				axis y line*=left,
				ylabel={\#Model Parameters, $\Pi$ (million)},
				xticklabel style={rotate=0},
				ymajorgrids = true,
				]
				\addplot+[draw=blue,fill=blue] coordinates {
					(E2E, 1.1556352) (WebNLG, 1.4926848) (Custom, 1.0691072)
				};
				\addplot+[draw=cyan,fill=cyan] coordinates {
					(E2E, 3.0490166) (WebNLG, 4.2618892) (Custom, 2.6160386)
				};
				\label{plot_y_mp_m4}
			\end{axis}
			
			\begin{axis}[
				ybar stacked,
				bar shift=10pt,
				xticklabels={},
				yticklabels={},
				]
				
				\addplot+[draw=teal,fill=teal] coordinates {
					(E2E, 1.1556352) (WebNLG, 1.4926848) (Custom, 1.0691072)
				};
				\addplot+[draw=green,fill=green] coordinates {
					(E2E, 3.5733074) (WebNLG, 4.7355812) (Custom, 3.3959944)
				};
				\label{plot_y_mp_m7}
			\end{axis}

			\begin{axis}[
				axis y line*=right,
				axis x line=none,
				yticklabel style={
                    /pgf/number format/fixed,
                    /pgf/number format/precision=5
                },
                scaled y ticks=true,
				ylabel={Vocab Size of Training split, $\left|\mathcal{V}\right|$},
				ymin=0, ymax=10000,
				legend style={at={(0.04,0.98)},anchor=north west}
				]
				
				\addplot+[mark=*,draw=brown,mark options={fill=yellow}]
				coordinates{				
					(E2E, 2620) (WebNLG, 8077) (Custom, 542)
				}; 
				\label{plot_y_vocab}
			
				\addlegendentry{\footnotesize $\left|\mathcal{V}\right|$}
				\addlegendimage{only marks,red,mark=square*}   \addlegendentry{\footnotesize $\Pi_{M_1}$}
				\addlegendimage{only marks,cyan,mark=square*}  \addlegendentry{\footnotesize $\Pi_{M_4}$}
				\addlegendimage{only marks,green,mark=square*} \addlegendentry{\footnotesize $\Pi_{M_7}$}
				
			\end{axis}
		\end{tikzpicture}
	}
	\caption{
	    \textbf{Effect of our architectural choices on model parameters and vocabulary}.
	    {\small
	    For all models, brighter shades denote decoder parameters, stacked on top of their encoder counterparts in darker shades.
	    }
	}
	\label{fig:modelParamsAndVocab}
\end{figure}

We investigate the effect of different methodical and architectural choices, and the results are tabulated in Table \ref{tab:ablationResults}.  From ablation results, we infer that using Bi-LSTM in the encoder gives an absolute improvement of 2.89\% and 3.77\% in BLEU score on the E2E and Custom datasets, respectively, as it generates efficient input representations than LSTM layer (used in $M_1$ encoder).  The impact of utilizing pre-trained (GloVe) embeddings can be seen from the fact that $M_3$ achieves 1.26\%, 3.23\%, 4.42\% and 0.81\% improvement in BLEU over $M_2$ on E2E, WebNLG: Seen, WebNLG: Unseen and Custom datasets, respectively.  Slightly better results with Bahdanau attention ($M_4$) are possible due to the additional weight matrices being trained as against Luong ($M'_4$) multiplicative style.  By contrast, higher BLEU scores of $M_6$ over $M_5$ for WebNLG and Custom explains the effectiveness of encoding message intent information as an additional input feature.  Furthermore, we observe an absolute improvement of 3.15\% and 2.41\% in BLEU and ROUGE-L scores (on Custom) by applying Beam Search (BS) decoding with beam width as 3.  Fig. \ref{fig:modelParamsAndVocab} depicts the variation in model parameters in $M_1$, $M_4$, and $M_7$, across three datasets (vocabulary unchanged across models). As we can observe, our heaviest model, $M_7$, requires only $\sim2\times$ the number of parameters of a basic encoder-decoder ($M_1$). Furthermore, this amounts to around 4 to 6 million parameters for E2E and WebNLG datasets that makes it much more suitable for on-device deployment compared to PLMs.\looseness=-1

\looseness=-1

\begin{table}[tbp]
	\caption{
	    Ablation study
	    ($\forall M_i$, ${}_tr_\mathcal{K} = 0$, ${}_er_\mathcal{K} = 0$)
	}
	\centering
	\resizebox{\linewidth}{!}{%
		\begin{tabular}{l c | c c | c c | c c}
			\toprule
			                               
			\multicolumn{2}{c|}{\multirow{4}{*}{\textbf{Model}}}       
			& \multicolumn{2}{c|}{\textbf{E2E}} 
			&   \multicolumn{2}{c|}{\textbf{WebNLG}}  
			&   \multicolumn{2}{c}{\textbf{Custom}}
			\\ \cline{3-8}
			
			& 
			& 
			& 
			& \multicolumn{2}{c|}{\textbf{BLEU}}
			& 
			& 
			\\
			                                                                               
			&
			& \textbf{BLEU} 
			& \textbf{ROUGE-L} 
			& \textbf{Seen}
			& \textbf{Unseen}
			& \textbf{BLEU}  
			& \textbf{ROUGE-L}
			\\
			
			&
			& \textbf{(\%)}
			& \textbf{(\%)} 
			& \textbf{(\%)} 
			& \textbf{(\%)} 
			& \textbf{(\%)} 
			& \textbf{(\%)}
			\\ \midrule \midrule

			Basic Encoder-Decoder    
			& $M_1$
			& 48.06     
			& 49.33              
			& 43.96          
			& 33.19         
			& 40.68         
			& 36.70                
			\\

			\hspace{0.3cm}+ BiLSTM (in $E_D$)                        
			& $M_2$        
			& 50.95           
			& 51.66              
			& 47.00     
			& 35.27          
			& 44.45         
			& 39.03           
			\\
			
			\hspace{0.6cm}+ GloVe embeddings  
			& $M_3$     
			& 52.21          
			& 53.38            
			& 50.23   
			& 39.69   
			& 45.26            
			& 42.96         
			\\
			
			{\color{gray}\hspace{0.9cm}+ Luong} 
			& {\color{gray}$M'_4$}  
			& {\color{gray}59.11}
			& {\color{gray}59.51}
			& {\color{gray} 56.77}     
			& {\color{gray} 43.52}   
			& {\color{gray}48.95}
			& {\color{gray}50.35} 
			\\
			
			\hspace{0.9cm}+ Bahdanau 
			& $M_4$            
			& 59.65             
			& 60.84          
			& 56.71 
			& 43.64  
			& 50.21         
			& 52.88        
			\\
			
			\hspace{1.2cm}+ Copy    
			& $M_5$           
			& 64.59          
			& 67.20           
			& 60.07 
			& 51.44 
			& 60.85         
			& 65.26         
			\\
			
			\hspace{1.5cm}+ Intent Input  
			& $M_6$         
			& {\color{gray}N/A}
			& {\color{gray}N/A}   
			& 62.89   
			& 51.29    
			& 65.79           
			& 72.04        
			\\
			
			{\color{blue}\textit{TrICy}$\rightarrow$}\hspace{0.2cm}\textbf{+ BS decoding}
			& \textbf{\textit{M$_\mathbf{7}$}}
			& \textbf{66.43}    
			& \textbf{70.14}     
			& \textbf{64.08}  
			& \textbf{52.35}  
			& \textbf{68.94}    
			& \textbf{74.45}     
			\\ 
			
			\bottomrule
		\end{tabular}
	}
	\label{tab:ablationResults}
\end{table}

\subsection{\texorpdfstring{$\mathcal{K}$}{K} triggers and the search for an optimum ratio, \texorpdfstring{${}_tr_\mathcal{K}^*$}{trK*}}\label{sec:triggerRatioExperiments}

\begin{figure*}[t]
	\centering
	\begin{minipage}{0.195\linewidth}
		\includegraphics[width=\linewidth, trim={.5cm .5cm .2cm .5cm}, clip]{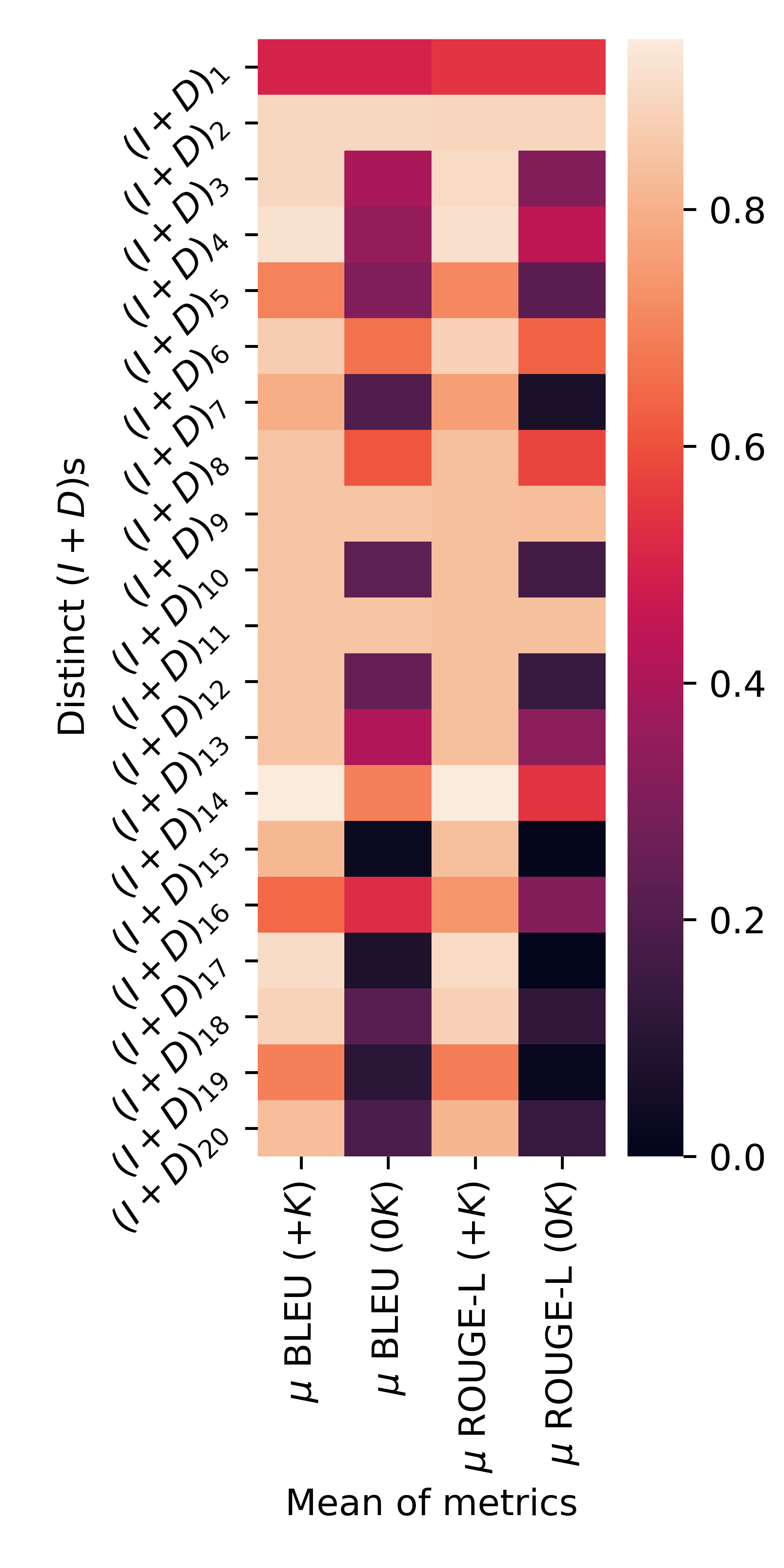}
		\caption*{\centering (a) ${}_tr_\mathcal{K} = 0.00$}
		\label{fig:metricHeatmaps_0.00}
	\end{minipage}
	\begin{minipage}{0.195\linewidth}
		\includegraphics[width=\linewidth, trim={.5cm .5cm .2cm .5cm}, clip]{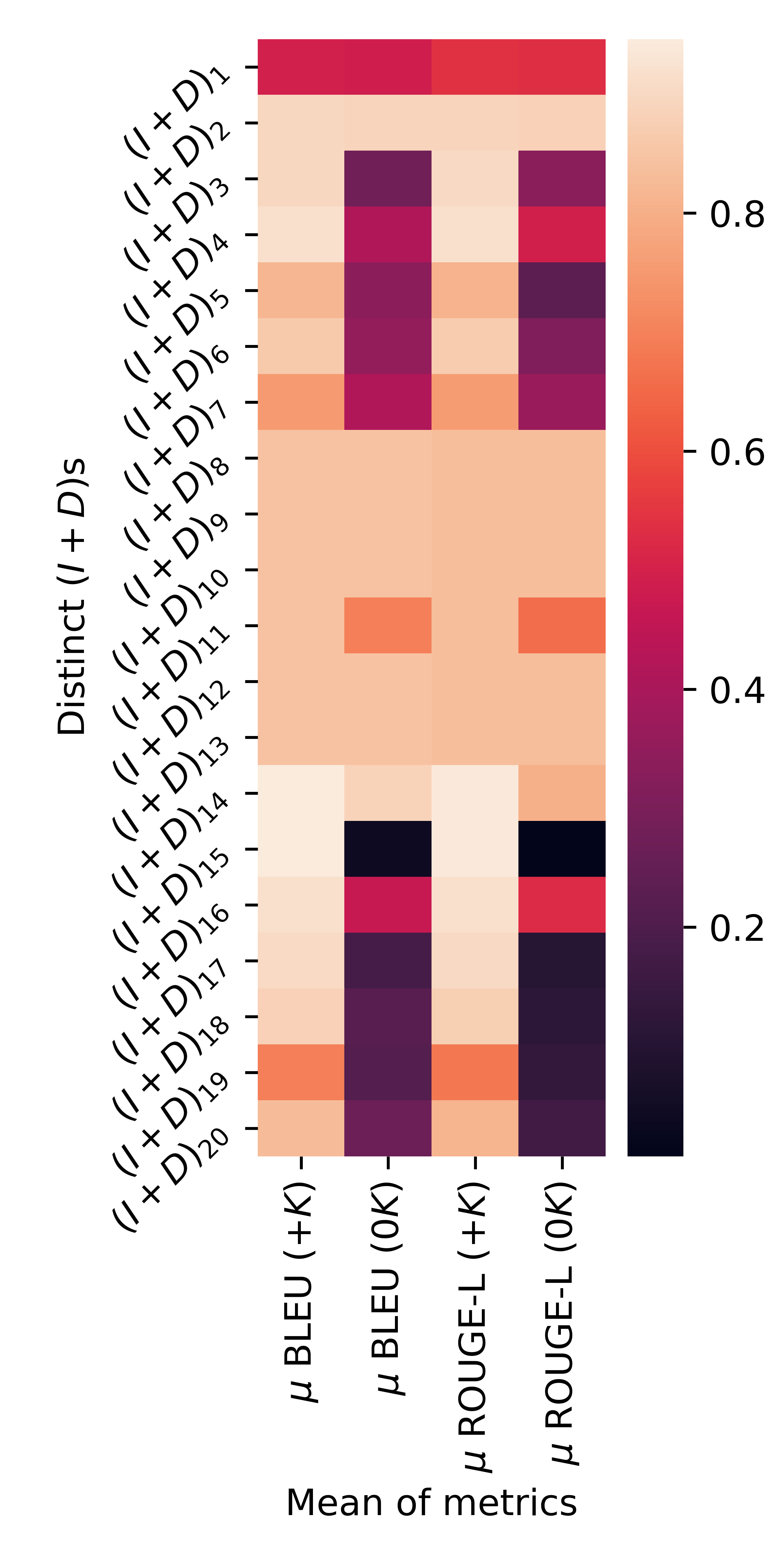}
		\caption*{\centering (b) ${}_tr_\mathcal{K} = 0.25$}
		\label{fig:metricHeatmaps_0.25}
	\end{minipage}
	\begin{minipage}{0.195\linewidth}
		\includegraphics[width=\linewidth, trim={.5cm .5cm .2cm .5cm}, clip]{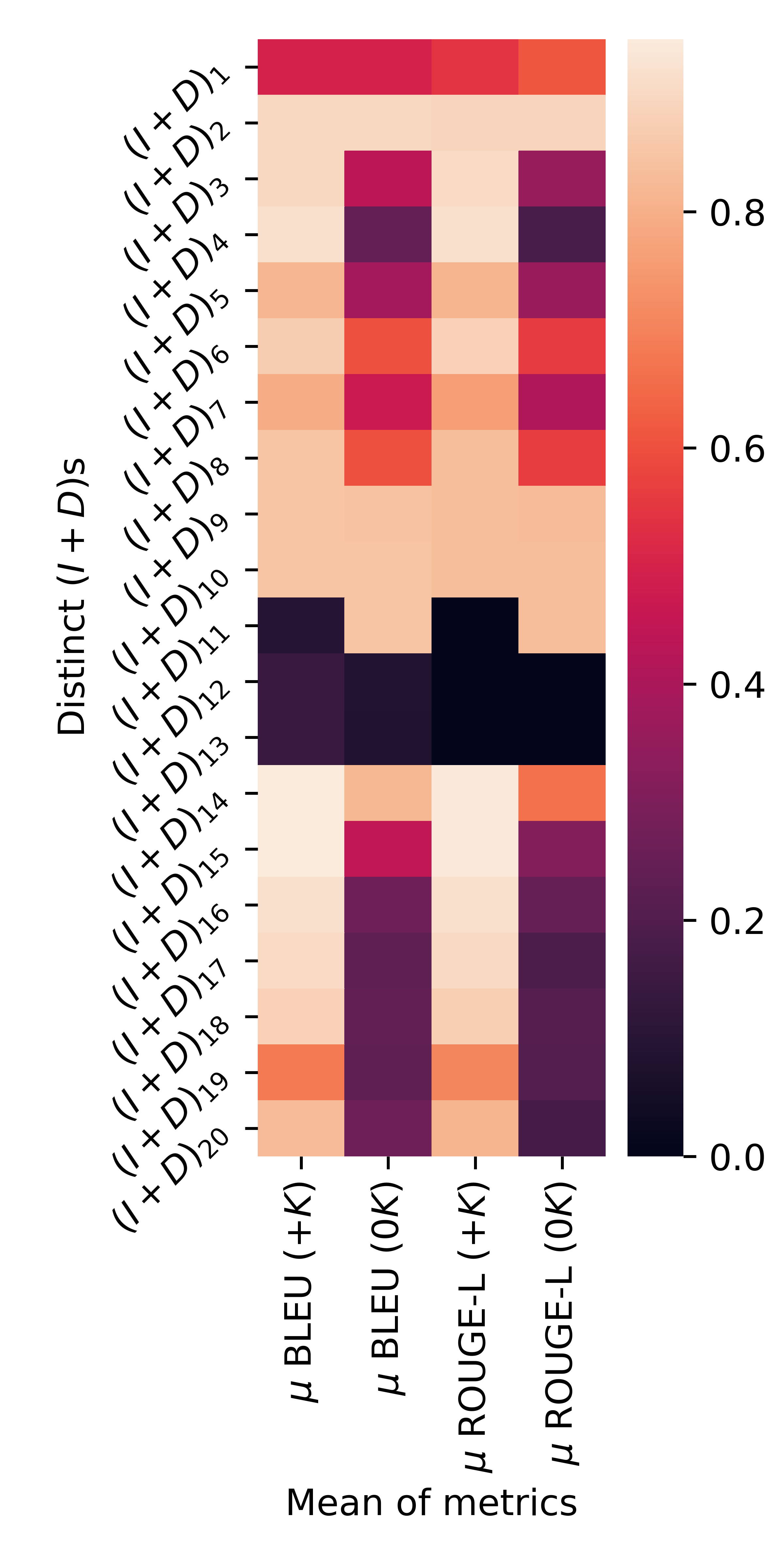}
		\caption*{\centering (c) ${}_tr_\mathcal{K} = 0.50$}
		\label{fig:metricHeatmaps_0.50}
	\end{minipage}
	\begin{minipage}{0.195\linewidth}
		\includegraphics[width=\linewidth, trim={.5cm .5cm .2cm .5cm}, clip]{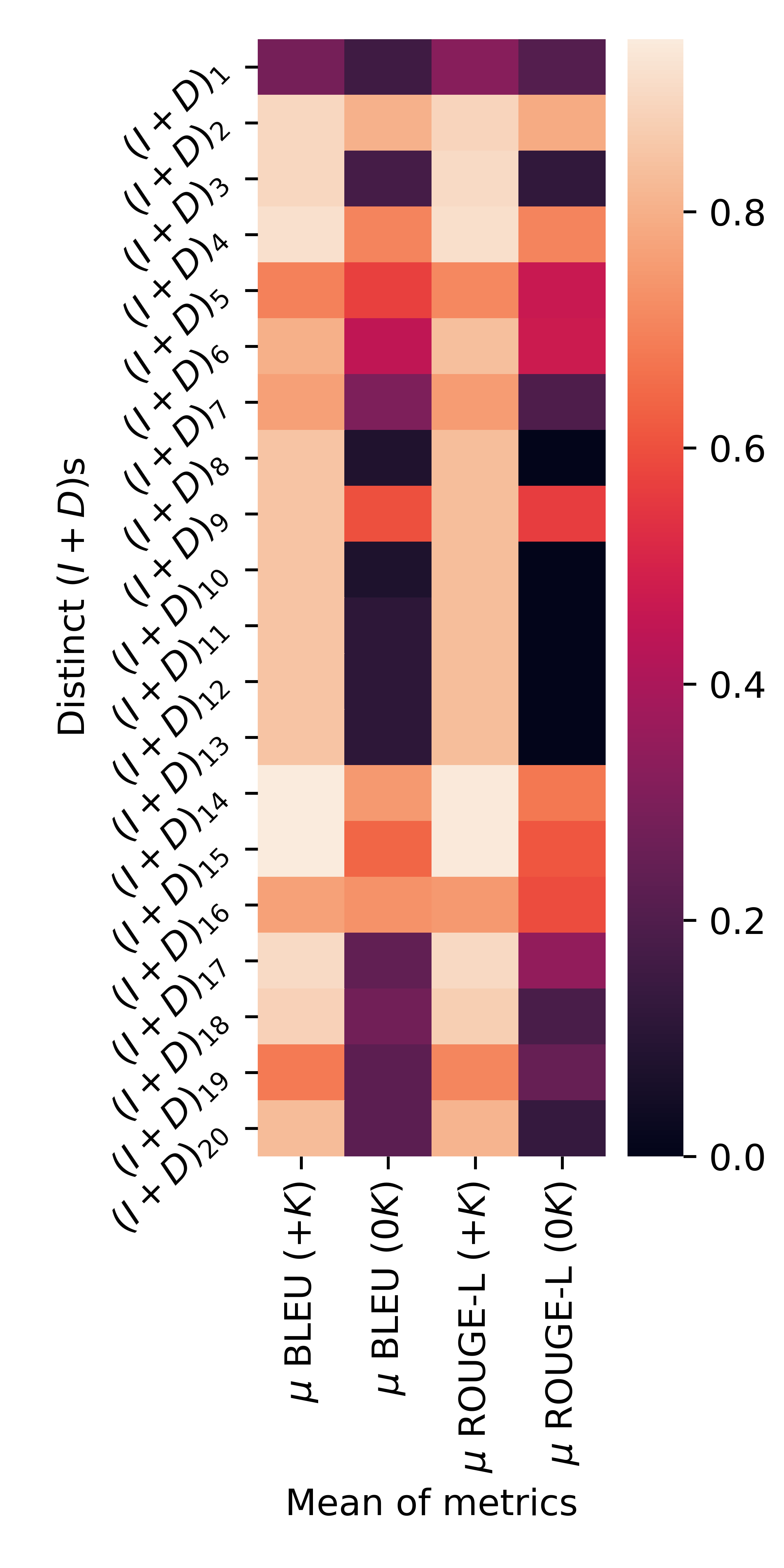}
		\caption*{\centering (d) ${}_tr_\mathcal{K} = 0.75$}
		\label{fig:metricHeatmaps_0.75}
	\end{minipage}
	\begin{minipage}{0.195\linewidth}
		\includegraphics[width=\linewidth, trim={.5cm .5cm .2cm .5cm}, clip]{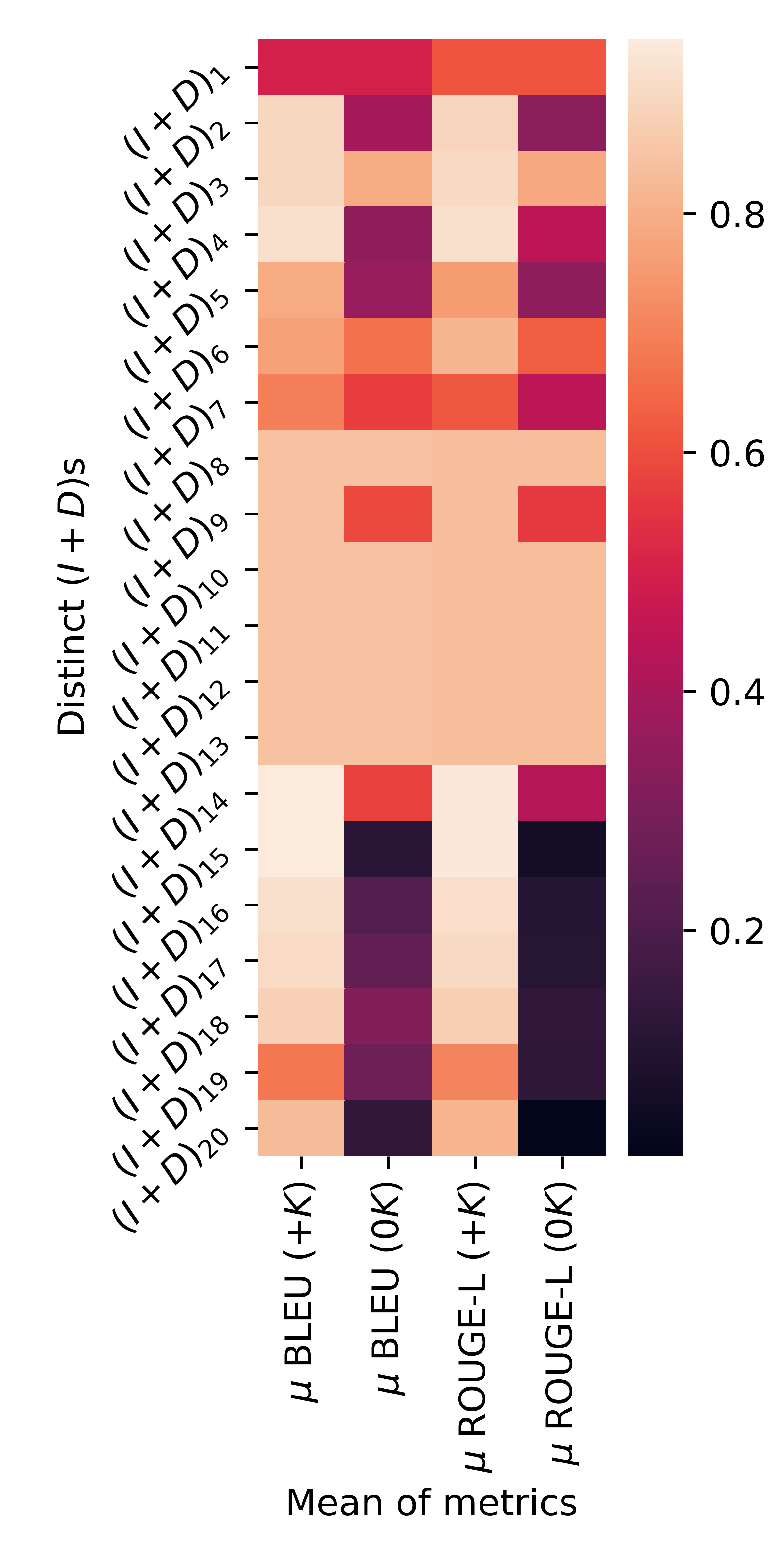}
		\caption*{\centering (e) ${}_tr_\mathcal{K} = 1.00$}
		\label{fig:metricHeatmaps_1.00}
	\end{minipage}
	\caption{\textbf{
	    Categorized evaluation of models trained using Custom dataset with selected ${}_tr_\mathcal{K}$ ratios:
	} {\small
	    The presence of trigger $\mathcal{K}$ in input significantly improves the generation of a directed output sequence. This improvement increases with ${}_tr_\mathcal{K}$. As a sidenote, from the description of $(I+D)$ aliases in Table \ref{tab:heatmapDescription}, it is observed that some intents, like \texttt{CALENDAR::ACT} $\in \mathcal{I}$, perform more consistently across all $\protect\overrightarrow{\mathcal{D}_c}$ field inputs than other intents.
	}}
	\label{fig:metricHeatmaps}
\end{figure*}

We conduct performance analysis on models trained with multiple configurations based on ${}_tr_\mathcal{K}$ introduced in Section \ref{sec:triggerInput}. The BLEU and ROUGE-L values observed during the evaluation of 5 notable model configurations, covering 6 of the intent classes in the Custom dataset, are presented in Fig. \ref{fig:metricHeatmaps}.  Here, in each configuration, the ${}_tr_\mathcal{K}$ is set to a constant value, and the mean ($\mu$) BLEU and ROUGE-L F1 scores are evaluated for two extreme scenarios of test set -- (i) where trigger word is absent in all evaluation samples, i.e., ${}_er_\mathcal{K} = 0.0$, denoted by $0K$, and (ii) where trigger word is present in all evaluation samples, i.e., ${}_er_\mathcal{K} = 1.0$, denoted by $+K$. We may observe that for all values of ${}_tr_\mathcal{K}$, providing a trigger input dramatically improves the BLEU and ROUGE-L scores.  Furthermore, the improvement due to the introduction of trigger increases as ${}_tr_\mathcal{K}$ increases.  The improvement with the $+K$ evaluation set is expected as intuitively increasing the presence of trigger input in training data trains the model to predict a directed text sequence that can be more aligned with the gold standard in the evaluation set.  However, increasing ${}_tr_\mathcal{K}$ may also lead to increased model dependence on the trigger that is correlated to frequent deterioration of metrics in $0K$ evaluation sets.  Therefore, we aim to find a trade-off between the two extremes for ${}_tr_\mathcal{K}$.\looseness=-1


\begin{figure}[htb]
	\centering
	
	\subfloat[\small Custom dataset]{%
	    \includegraphics[width=0.79\linewidth, trim={.8cm .8cm .4cm .7cm}, clip]{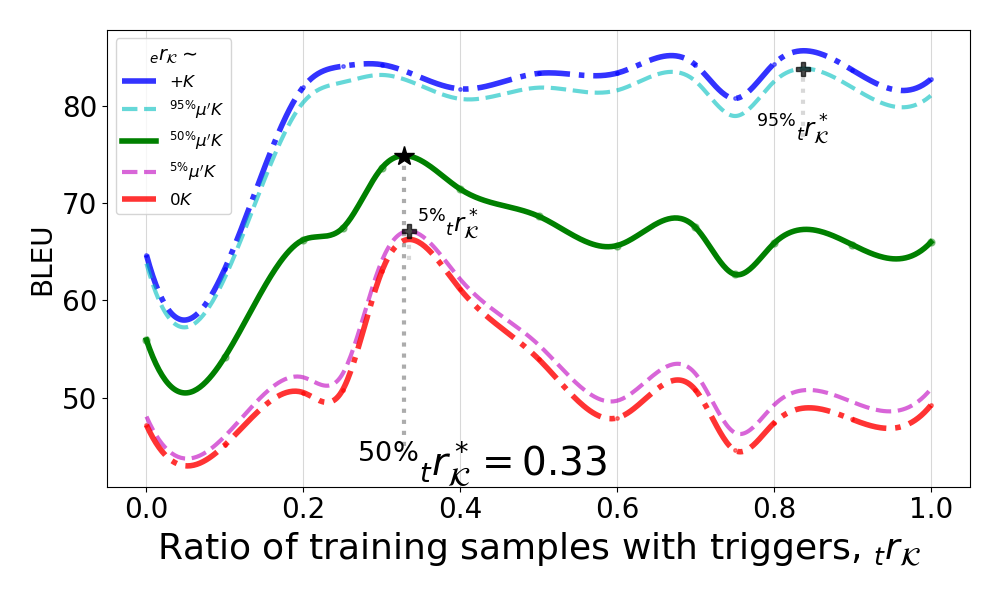}
	     \label{fig:lineGraphBleuSynthetic}
	}
	
	\subfloat[\small E2E dataset]{%
	    \includegraphics[width=0.79\linewidth, trim={.8cm .8cm .4cm .7cm}, clip]{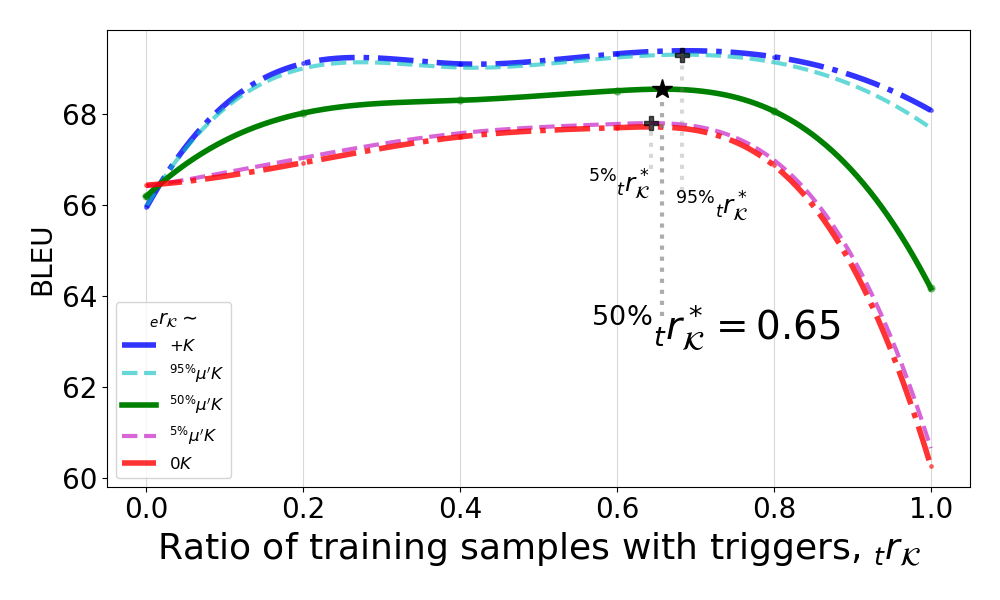}
	    \label{fig:lineGraphBleuE2E}
	}

	\caption{\textbf{
	    Determination of ${}_tr_\mathcal{K}^*$:
	} {\small
	    For models trained with varying ${}_tr_\mathcal{K}$ ratios with evaluation sets -- (i) $0K$ (${}_er_\mathcal{K} = 0.0$), and (ii) $+K$ (${}_er_\mathcal{K} = 1.0$), the weighted mean graphs of (i) and (ii) are denoted by ${}^{w\%}\mu'K$, for heuristic weight $w\%$.
	}}
	\label{fig:lineGraphBleu}
\end{figure}

To identify the optimal value for ${}_tr_\mathcal{K}$, namely ${}_tr_\mathcal{K}^*$, we analyze the BLEU scores aggregated over all test samples for the $0K$ and $+K$ evaluation sets as shown in Fig. \ref{fig:lineGraphBleu}.  An interesting observation is that in Fig. \ref{fig:lineGraphBleuSynthetic}, even when ${}_tr_\mathcal{K}$ is set to 0, i.e., when the model has been trained with no training sample with trigger input, providing trigger in test input consistently improves the model performance.  We can attribute this improvement to an implicit internal transfer of the learned value embeddings for $\overrightarrow{\mathcal{D}_c}$ to the otherwise \texttt{<SOS>} token for trigger $\mathcal{K}$.\looseness=-1

In a real-world scenario, it is likely that input samples do not always have any predetermined trigger. Yet, there are times when the incorporation of such a trigger may be relevant and can help generate a more desired output sequence. By taking a heuristic-based weight $w\%$ (how often a trigger is likely to be present) and the corresponding weighted mean of the two curves, denoted as ${}^{w\%}\mu'K$. Then, by locating its peak on the Y-axis, we determine the optimum ratio ${}^{w\%}{}_tr_\mathcal{K}^*$ using the corresponding value on the X-axis. Thus,\looseness=-1
\begin{equation}
    \begin{aligned}
        \forall{w \in [0,100]}, \Hquad && {}^{w\%}{}_tr_\mathcal{K}^* = \argmax_{{}_tr_\mathcal{K}} \Big( {}^{w\%}\mu'K \Big)
    \end{aligned}
    \label{eq:optimumTriggerRatio_first}
\end{equation}
where,
\begin{equation}
    \small
    \begin{aligned}
        & {}^{w\%}\mu'K & = \Hquad & \frac{w}{100} * \big(\left(+K\right)\big) & + & \left(1 - \frac{w}{100}\right) * \big(\left(0K\right)\big)\\
        &  & = \Hquad & \frac{w}{100} * \left(\metric_{{}_er_\mathcal{K} = 1}\right) & + & \left(1 - \frac{w}{100}\right) * \left(\metric_{{}_er_\mathcal{K} = 0}\right)\\
        \therefore \Hquad & {}^{w\%}\mu'K & \approx \Hquad & \metric_{{}_er_\mathcal{K} = w\%}
    \end{aligned}
    \label{eq:optimumTriggerRatio_second_a}
\end{equation}
From Equations \eqref{eq:optimumTriggerRatio_first} and \eqref{eq:optimumTriggerRatio_second_a}, we can observe that
\begin{equation}
    \begin{aligned}
        {}^{w\%}{}_tr_\mathcal{K}^* & = \argmax_{{}_tr_\mathcal{K}} \Bigg( \metric_{{}_er_\mathcal{K} = w\%} \Bigg)
        \label{eq:optimumTriggerRatio}
    \end{aligned}
\end{equation}

For any intended application, the heuristic, $w$, can be decided based on the existing domain knowledge or big data analysis of market behavior. In our experiments, we use 50\% as the heuristic weight in Equation \eqref{eq:optimumTriggerRatio} for averaging the two curves, and as depicted in Fig. \ref{fig:lineGraphBleu}, we compute ${}_tr_\mathcal{K}^*$ to be 0.33 for Custom dataset and 0.65 for E2E dataset. Similarly, we compute ${}_tr_\mathcal{K}^*$ to be 0.24 for WebNLG dataset.\looseness=-1

\subsection{Discussion and Applications}\label{sec:discussionandapplication}

\subsubsection{Controlled investigation into the effect of \texorpdfstring{$\mathcal{K}$}{K}, \texorpdfstring{$\mathcal{I}$}{I}, \texorpdfstring{$\protect\overrightarrow{\mathcal{D}_c}$}{Dc}} \label{sec:effectOfKID}
Table \ref{tab:qualitativeResults} illustrates sample outputs from TrICy when trained on the Custom dataset. We see that by simply varying input for $\mathcal{I}$ (like \texttt{CONTACT::SHARE} $\rightarrow$ \texttt{CONTACT::ACT}), while retaining the same $\protect\overrightarrow{\mathcal{D}_c}$ and $\mathcal{K}$, output $\protect\overrightarrow{\mathcal{S}_{g,c}}$ is either a human-readable text sequence or a markup sequence that can enable rendering of intelligent action prompts (like \textit{``call a person''}, \textit{``edit a scheduled meeting''}, etc.) on edge devices. Moreover, the presence of trigger $\mathcal{K}$ leads to variations in output generations, as evident from the tabulated examples.

\begin{table*}[tb]
	\caption{{ 
	    Sample of a few example text sequences generated by TrICy (${}_tr_\mathcal{K}^*$ = 0.33) on Custom dataset
	}}
	\centering
	\resizebox{0.92\linewidth}{!}{%
		\begin{tabular}{c c p{0.2\linewidth} p{0.6\linewidth}}
			\toprule
			
			\multicolumn{3}{c}{\textbf{Input}} 
			& \multicolumn{1}{c}{\textbf{Output}}    
			\\ \cline{1-3}
			\\[-0.9em]
			
			\multicolumn{1}{c}{\textbf{$\mathcal{K}$}} & \multicolumn{1}{c}{\textbf{$\mathcal{I}$}} 
			& \multicolumn{1}{c}{\textbf{$\overrightarrow{\mathcal{D}_c}$}}  
			& \multicolumn{1}{c}{\textbf{$\overrightarrow{\mathcal{S}_{g,c}}$}}   
			\\ \midrule \midrule
			         
			``\texttt{<SOS>}''
			& \texttt{CONTACT::SHARE}    
			& \{\texttt{Name}: ``{\color{teal} John}'', \newline \texttt{Phone}: ``{\color{blue} 9867452301}''\}
			& \textit{You can call {\color{teal} John} at {\color{blue} 9867452301}}
			\\ \midrule
			         
			``\texttt{<SOS>}'' 
			& \texttt{CONTACT::ACT}   
			& \{\texttt{Name}: ``{\color{teal} John}'', \newline \texttt{Phone}: ``{\color{blue} 9867452301}''\}
			& \textit{$\langle call \rangle \langle name \rangle \text{\color{teal} John} \langle /name \rangle \langle number \rangle \text{\color{blue} 9867452301} \langle /number \rangle \langle /call \rangle$} $\implies \vcenter{\hbox{\includegraphics[height=3\fontcharht\font`\B]{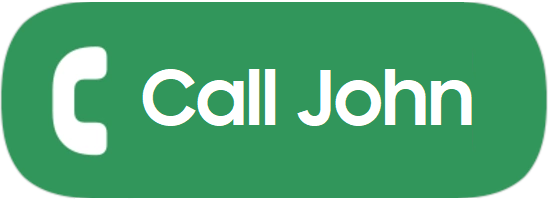}}}$
			\\ \midrule
			         
			``\texttt{<SOS>}''
			& \texttt{CALENDAR::ACT}
			& \{\texttt{ID}: ``{\color{orange} 1325406}'', \newline \texttt{Title}: ``{\color{violet} Staff Call}'', \newline \texttt{TimeStart}: ``{\color{olive} 2 PM}'', \newline \texttt{TimeEnd}: ``{\color{magenta} 3 PM}'', $\cdots$ \}
			& \textit{$\langle event\_edit \rangle \langle id \rangle \text{\color{orange} 1325406} \langle /id \rangle \langle title \rangle \text{\color{violet} Staff Call} \langle /title \rangle \newline \cdots \langle /event\_edit \rangle$} \newline $\implies \vcenter{\hbox{\includegraphics[height=4\fontcharht\font`\B]{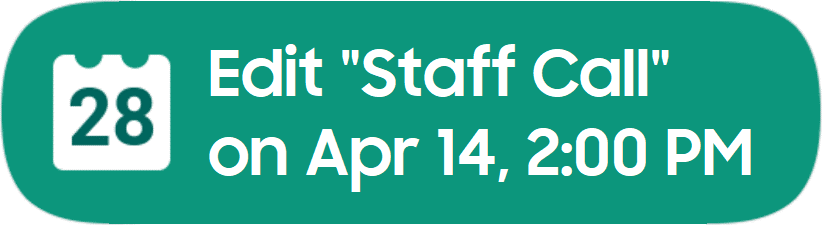}}}$
			\\ \midrule
			         
			``\texttt{<SOS>}''
			& \texttt{CALENDAR::SHARE}
			& \{\texttt{ID}: ``{\color{orange} 1325406}'', \newline \texttt{Title}: ``{\color{violet} Staff Call}'', \newline \texttt{TimeStart}: ``{\color{olive} 2 PM}'', \newline \texttt{TimeEnd}: ``{\color{magenta} 3 PM}'', $\cdots$ \} 
			& \textit{{\color{violet} Staff Call} is scheduled on {\color{darkgray} April 14} from {\color{olive} 2 PM} to {\color{magenta} 3 PM}} 
			\\ \midrule
			         
			``\underline{meeting}'' 
			& \texttt{CALENDAR::SHARE}
			& \{\texttt{ID}: ``{\color{orange} 1325406}'', \newline \texttt{Title}: ``{\color{violet} Staff Call}'', \newline \texttt{TimeStart}: ``{\color{olive} 2 PM}'', \newline \texttt{TimeEnd}: ``{\color{magenta} 3 PM}'', $\cdots$ \}
			& \textit{\underline{Meeting} for {\color{violet} Staff Call} is from {\color{olive} 2 PM} to {\color{magenta} 3 PM} on {\color{darkgray} April 14}}
			\\ \midrule
			         
			``\texttt{<SOS>}'' 
			& \texttt{LOCATION::SHARE}
			& \{\texttt{Location}: \newline ``{\color{teal} Central Park, NYC}''\}
			& \textit{I am at {\color{teal} Central Park, NYC}}
			\\ \midrule
			         
			``\texttt{<SOS>}''
			& \texttt{OCCASION::SHARE}
			& \{\texttt{Name}: ``{\color{teal} Sarah}'', \newline \texttt{Occasion}: ``{\color{cyan} New Year}''\}
			& \textit{Happy {\color{cyan} New Year}}, {\color{teal} Sarah}
			\\ \midrule
			         
			``\underline{wish}''
			& \texttt{OCCASION::SHARE}
			& \{\texttt{Name}: ``{\color{teal} Sarah}'', \newline \texttt{Occasion}: ``{\color{cyan} New Year}''\}
			& \textit{\underline{Wish} you a very happy {\color{cyan} New Year}, {\color{teal} Sarah}}
			\\ \midrule
			         
			\bottomrule
		\end{tabular}
	}
	\label{tab:qualitativeResults}
\end{table*}

\subsubsection{Insights from User Trial} \label{sec:userTrial}
To evaluate the end-user perspective of TrICy-generated output sequences, we conduct a closed user trial using a custom Android app with $N_u (=48)$ users. After going through a list of illustrated instructions and consenting to participation, each user is presented with $N_s (=100)$ test samples. In each of these samples, the intent (like \texttt{CONTACT::ACT} or \texttt{CALENDAR::SHARE}) is pre-selected at random by the app, and a relevant information sample (like details of a contact or calendar event) is extracted from the existing user data. These are then passed to the on-device TrICy engine to generate a response sequence. The user may accept the default suggestion (event $= A_d$) or add a trigger text to request an alternative sequence. Thereafter, the user can either accept the alternative sequence (event $= A_a$) or reject it (event $= A_r = \overline{A_d} \land \overline{A_a} = \overline{A_d \lor A_a}$). We then aggregate these results across various intents and users and compute the metrics, $P \left( A_d \right)$ and $P \left( A_a \mid \overline{A_d} \right)$:

\begin{equation}
    \small
    \begin{aligned}
            P \left( A_d \right) &= \frac{\left|\text{User accepts default suggestion}\right|} {N_u \times N_s}\\
            P \left( A_a \mid \overline{A_d} \right) &= \frac{\left|\text{User accepts alternative suggestion}\right|} {\left|\overline{\text{User accepts default suggestion}}\right|}
    \end{aligned}
\end{equation}

From our user trial, we compute $P \left( A_d \right)$ and $P \left( A_a \mid \overline{A_d} \right)$ to be 0.6721 and 0.5648 respectively. This implies that the users would prefer to use the default suggestions generated by TrICy in $\sim67\%$ of cases, and out of the remaining $\sim33\%$, they can derive a suitable alternative suggestion by providing a trigger text for $\sim56\%$ of such cases. This indicates the overall likelihood of usefulness of TrICy suggestions to be $0.6721 + 0.5648 \times ( 1 - 0.6721 ) = 85.73\%$.

\subsubsection{Advantage over templates}\label{sec:advantageOverTemplates}

Limitations of na\"ive template-based approach include:

\begin{enumerate}[label=(\roman*)]
    \item Templates are restrictive as this necessitates manual curation of templates for each new intent and field combination.
    
    \item It is cumbersome to extend post-deployment using additions in training samples as that requires continuous manual identification of patterns and template updates.

    \item Templates do not offer the flexibility of personalization beyond a random selection or keyword-based lookup from a preloaded set of statements.\looseness=-1
\end{enumerate}

To address these shortcomings, we leverage the optional ``trigger'' feature, introduced in Section \ref{sec:triggerInput} and further explored in Section \ref{sec:triggerRatioExperiments}, that provides multiple degrees of personalization. A trigger can determine a more directed output personalized to the user and context. 
It also supports continuous training, making it preferable for techniques like Federated Learning.

\subsubsection{Potential Applications}\label{sec:potentialApplications} The competitive accuracy, tiny memory footprint, and low inference time of TrICy enable applications like:   
\begin{enumerate}[label=(\roman*)]
    \item Contextual recommendations: As illustrated in Fig. \ref{fig:smartComposeSimple}, by utilizing received message(s) intent and optional input trigger, data is retrieved from the user's application database to provide intelligent responses on the toolbar of mobile's soft keyboard. Moreover, adaptive phrase suggestions can also be provided based on the user's typed text. 

    \item Synthetic Data Creation: Collecting real data to train machine learning models is not always feasible due to cost, privacy, and availability factors \cite{syntheticdata}. Hence, synthetic data is a good alternative for training data-hungry neural models. TrICy can be leveraged to generate textual data that can be used to train conversational AI models and intelligent voice assistants, enhance text-to-speech systems or generate photo descriptions from the given image tags.\looseness=-1
    
    \item Meaningful Screen Reader: With a little modification, the proposal can be generalized and applied to other data-to-text tasks that require inference on edge devices like Meaningful Screen Reader. Such a module can aid users with accessibility needs by converting data points appearing on the screen to more natural and semantically meaningful text sequences. As depicted in Table \ref{tab:qualitativeResults}, based on the relative importance of each data input field, markup tags can also be generated in such text output to signal intonations for a downstream text-to-speech (TTS) engine.
\end{enumerate}

\section{Conclusion}\label{sec:conclusion}

We propose a novel D2T generation framework, TrICy, which utilizes the intent information and optional triggers for generating diverse and contextual outputs. We evaluate the effectiveness of attention+copy and intent through standard objective metrics against two public and one Custom datasets.
Furthermore, we explore the effect of proposed architecture components through an ablation study in Section \ref{sec:ablation}. Our extensive experiments on trigger-guided training in Section \ref{sec:triggerRatioExperiments} present a method to compute the optimal trigger ratio for any training set.
Leveraging message intent and triggers, our best TrICy model achieves the new SOTA performance with lowest memory footprint.\looseness=-1

Our approach is more flexible and scalable than handcrafting a template for each generation scenario, as explained in Section \ref{sec:advantageOverTemplates}. Furthermore, our model can generate adaptive sequences that enable various real-world applications (see Section \ref{sec:potentialApplications} for details). In future, we intend to investigate support for phonetic scripts and evaluate the efficacy of our approach on code-switched languages.\looseness=-1

\section*{Limitations}\label{sec:limitations}
Our detailed experiments focus on English datasets and do not evaluate other languages. Hallucination is a crucial problem in natural language generation where the generated content is unfaithful to the provided source content. Data-to-text generation is particularly prone to this \cite{Ji2022SurveyOH}. Quantifying the effect of such issues concerning triggers is not discussed in this paper, and we intend to explore this in future work.\looseness=-1

\section*{Ethical Considerations}\label{sec:ethicalconsiderations}
In this paper, we propose to generate syntactically correct sequences based on these inputs: data record, intent, and optional trigger. The flexibility offered by trigger input may lead to increased vulnerability to adversarial attacks. However, we expect this to be minimizable by limiting the set of acceptable triggers. For all contributors taking part in data curation and user trial, the due consent process has been undertaken for the purpose of this work.\looseness=-1

\bibliographystyle{IEEEtran}
\bibliography{anthology,TrICy}

\newpage
\begin{IEEEbiography}[{\includegraphics[width=1in,height=1.25in,clip,keepaspectratio]{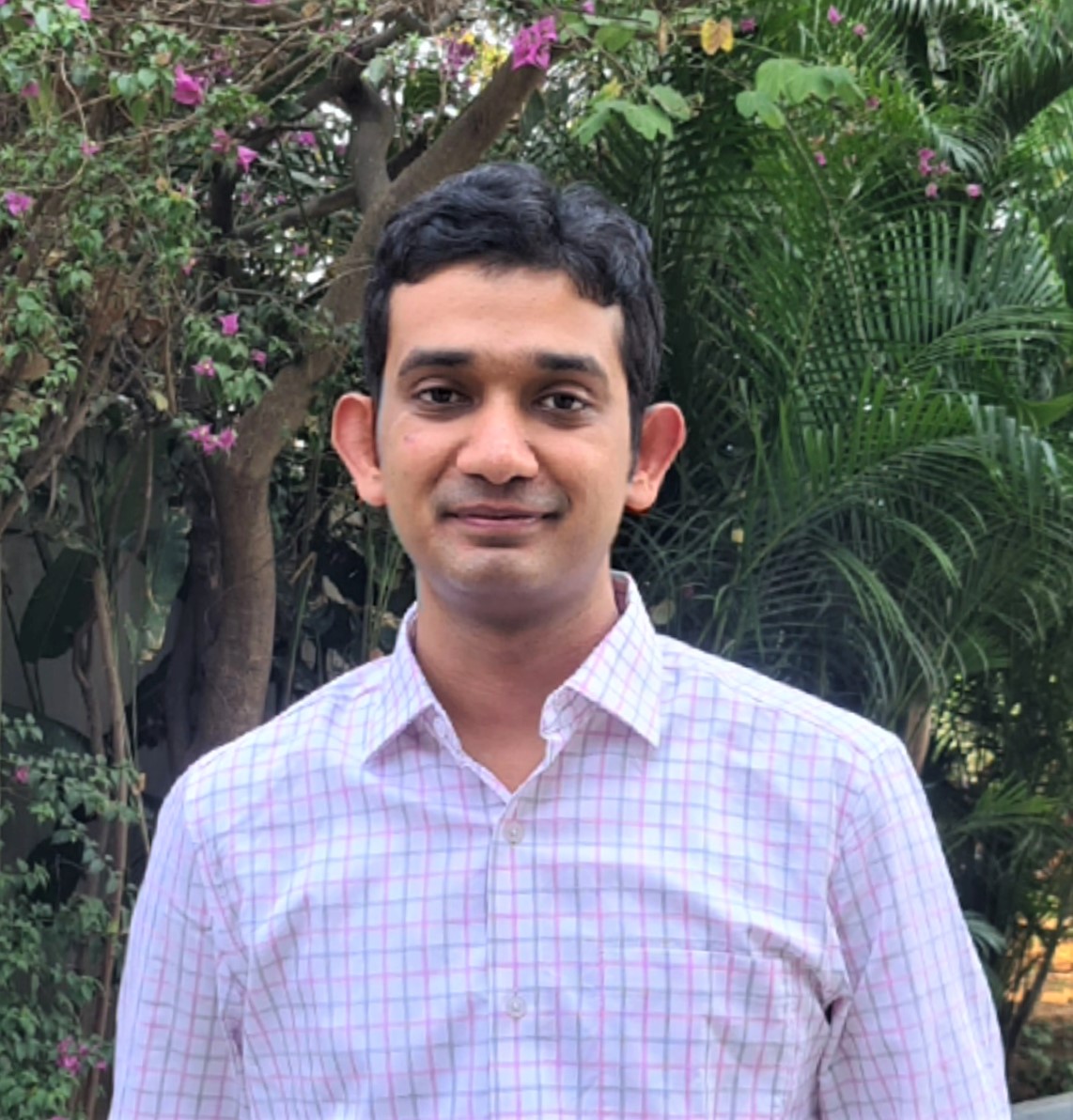}}]{Vibhav Agarwal}
is currently working as a Senior Chief Engineer at Samsung R\&D Institute Bangalore. He received his B.E. (Hons) in Manufacturing Engineering and M.Sc. (Tech) in Information Systems from Birla Institute of Technology and Science, Pilani in 2015. His current research interests include on-device AI, natural language processing, generative AI, affective computing, and computer vision. He has filed 15 patents and published 8 papers in these domains, leading to one Samsung best paper award and 2 nominations. He has actively contributed in development and successful commercialization of solutions for Samsung Galaxy devices, like transliteration, key area correction in the soft keyboard, emoji search and prediction, and smart reply. He was a recipient of Samsung Annual Award (Young Achiever of the Year)  for his technical and non-technical research-to-market contributions.
\end{IEEEbiography}

\vspace{-5 mm}

\begin{IEEEbiography}[{\includegraphics[width=1in,height=1.25in,clip,keepaspectratio]{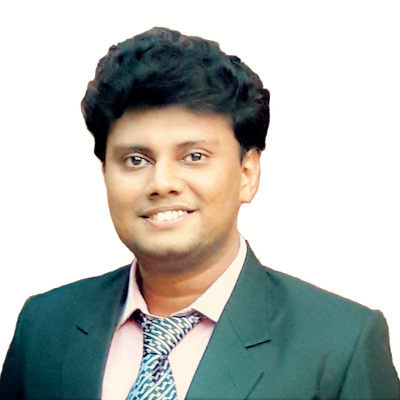}}]{Sourav Ghosh}
is currently working as a Chief Engineer at Samsung R\&D Institute Bangalore. He received his B.Tech degree in Information Technology from the West Bengal University of Technology, Kolkata, India, in 2014 and his M.Tech degree in Computer Science from the Indian Institute of Technology Kharagpur, Kharagpur, India, in 2017. His current research interests include on-device AI, natural language processing, generative AI, computer vision. He has published over 15 research papers in peer-reviewed forums, leading to 1 best paper award and 2 nominations, and has filed over 7 patents. He has served as a reviewer and program committee member for over 10 international conferences, including ACL, NeurIPS, EMNLP, ICCV, ACL-IJCNLP, NAACL-HLT, ICON, CSAE, and has reviewed over 40 manuscripts. He has mentored 35 teams as part of PRISM, an industry-academia collaboration and has served as a judge at Smart India Hackathon, hosted by the Ministry of Education, Government of India, for two consecutive years. Earlier in his career, he has worked with Tata Consultancy Services and as a Microsoft Student Partner. He has played an active role in the development and successful commercialization of mass-market solutions, like multilingual input method (Alive Intelligence), on-device Indic language models, as well as B2B solutions for airline digital marketing, defence communications, and in-cabin experiences. He has received Samsung Excellence Award - Innovator (Publications), Samsung Citizen Awards (x2), TCS LIREL Honor Rolls, Star of LG, IBM TGMC Winners Award.
\end{IEEEbiography}

\vspace{-5 mm}

\begin{IEEEbiography}[{\includegraphics[width=1in,height=1.25in,clip,keepaspectratio]{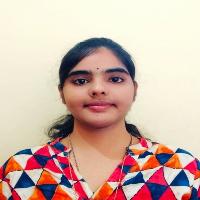}}]{Harichandana B S S}
is a Senior Engineer at Samsung R\&D Institute, Bangalore. She received the B.Tech degree
 in Information Technology from the National Institute of Technology, Surathkal, India in 2019. She is currently
 pursuing her M.Tech in Artificial Intelligence at the Indian Institute of Science, Bangalore. 
 She focuses on research in various domains like natural language processing, computer vision and security.  
 She has published over 8 papers receiving 3 best paper awards and one nomination and has filed over 4 patents.
She also served as a reviewer in AISTATS for over 2 manuscripts. She is currently working on non-intrusive on-device affective computing
and various solutions to enhance user experience.
\end{IEEEbiography}

\vspace{-5 mm}

\begin{IEEEbiography}[{\includegraphics[width=1in,height=1.25in,clip,keepaspectratio]{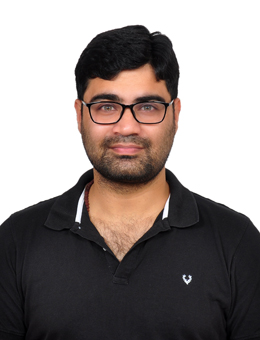}}]{Himanshu Arora}
is a Senior Chief Engineer in Samsung R\&D Institute Bangalore. He received his B.Tech in Information Technology from Malaviya National Institute of Technology, Jaipur in 2009. His work includes on-device AI, AI framework/platform. His current research interests include on-device AI, federated learning. He has filed 11 patents and published 3 papers. He has played an active role in the development and commercialization of solutions like word prediction model and emoji prediction, search in Samsung keyboard.
\end{IEEEbiography}

\vspace{-5 mm}

\begin{IEEEbiography}[{\includegraphics[width=1in,height=1.25in,clip,keepaspectratio,trim=1.5px 1.5px 1.5px 1.5px]{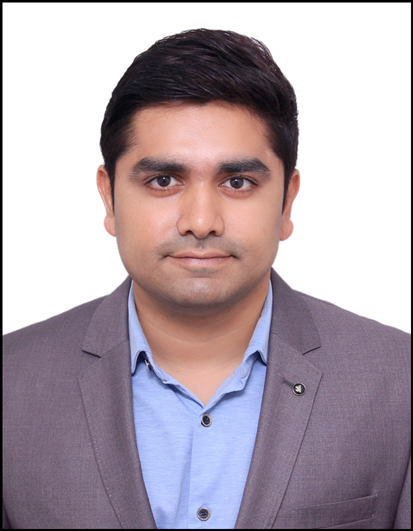}}]{Barath Raj Kandur Raja}
is a Principal Architect in Samsung R\&D Institute Bangalore and focuses
on ideating and commercializing impactful solutions, that influence end user experiences in smartphones and wearables. His work includes bringing AI powered experiences, on-device AI, AI framework/platform, affective computing, natural language processing/understanding, mobile application and framework domains. He has filed 55 patents and published 9 papers in these areas.
He received the Bachelor of Engineering in Computer Science and Engineering from Anna University, Chennai, Post Graduate Diploma in Data Science from the International Institute of Information Technology, Bangalore and Master of Science in Data Science from Liverpool John Moores University. He is currently pursuing a Global Doctor of Business Administration from the Swiss School of Business and Management, Geneva.
\end{IEEEbiography}

\vspace{10mm}

\appendix
\subsection{Configuration for experiments with Llama 2}\label{sec:LlamaFineTuningParameters}
For LLM fine-tuning, we follow the mechanism described by Labonne \cite{llama-code-tutorial}. Furthermore, we present the hyperparameters used to fine-tune Llama~2 7B model in Table~\ref{tab:llamaParams}. Models are fine-tuned on E2E NLG and WebNLG datasets for experiments described in Section~\ref{sec:comparisonWithSOTA}. For each dataset, we design text-completion prompts that conform to the format: \textit{``\texttt{<TAG\_START>} Question:} $f_1$[$v_1$], $f_2$[$v_2$], $\cdots$, $f_N$[$v_N$] \textit{\texttt{<TAG\_MID>} Answer: \texttt{<RESPONSE>} \texttt{<TAG\_END>}''}, where $f_i$ and $v_i$ refers to the $i^{\text{th}}$ field and value of the data sample respectively, and \textit{``\texttt{<RESPONSE>}''} indicates the reference sentence during fine-tuning and is generated during inference. The first occurrence of \textit{``\texttt{<TAG\_END>}''} in the generated text is used to demarcate the end of response to avoid hallucination.

\subsection{Custom dataset samples}\label{sec:CustomDatasetSamples}
Table~\ref{tab:customExamples} depicts a few examples from the Custom dataset for different $(I+D)$s illustrated in Table~\ref{tab:heatmapDescription}.

\begin{table}[bp]
	\caption{\small Hyperparameters for fine-tuning Llama 2 (used in evaluation in Table \ref{tab:sotaEvalWebNLG})}
	\centering
	\resizebox{0.9\linewidth}{!}{%
		\begin{tabular}{c | r}
			\toprule
			
			\textbf{Configuration} & \textbf{Value} \\
			
			\midrule\midrule
            \multicolumn{2}{c}{\textit{QLoRA parameters}} \\ \midrule
            LoRA attention dimension & 64 \\
            Alpha parameter for LoRA scaling & 16 \\
            Dropout probability for LoRA layers & 0.1 \\
            
            \midrule
            \multicolumn{2}{c}{\textit{bitsandbytes parameters}} \\ \midrule
            4-bit precision base model loading & True \\
            Compute dtype for 4-bit base models & float16 \\
            Quantization type (fp4 or nf4) & nf4 \\
            Nested quantization for 4-bit base models & False \\
            
            \midrule
            \multicolumn{2}{c}{\textit{Training arguments parameters}} \\ \midrule
            \#Training epochs & 1 \\
            fp16/bf16 training & fp16 \\
            Batch size per GPU for training & 4 \\
            Batch size per GPU for evaluation & 4 \\
            \#Update steps to accumulate the gradients for & 2 \\
            Gradient checkpointing & True \\
            Maximum gradient normal (gradient clipping) & 0.3 \\
            Initial learning rate (AdamW optimizer) & $2 \times 10^{-4}$ \\
            Weight decay for all layers except bias/LayerNorm & 0.001 \\
            Optimizer to use & ``paged\_adamw\_32bit'' \\
            Learning rate schedule & ``constant'' \\
            Ratio of steps for linear warmup (0 to learning rate) & 0.03 \\
            Group sequences into batches with same length & True \\
            
            \bottomrule
			
		\end{tabular}
	}
	\label{tab:llamaParams}
\end{table}

\begin{table*}[hb]
	\caption{
	    {\normalfont 
	        Sample of Custom dataset, one item for each $(I+D)$ indicated in Table \ref{tab:heatmapDescription}
	    }
	}
	\centering
	\resizebox{.765\linewidth}{!}{%
    	\begin{adjustbox}{angle=90}
    		\begin{tabular}{c | p{0.2\linewidth} | p{0.2\linewidth} | p{0.52\linewidth}}
    			\toprule
    			$\mathcal{I}$
    			& $X_f$
    			& $X_v$
    			& Gold Reference
    			\\ \midrule \midrule
    			\multirow{3}{*}{CONTACT::ACT}
    			
    			& Contact, Email, Name  
    		    & 9078563412, noah@gmail.com, William 
    		    & $\langle\text{\texttt{T\_CONTACT\_BOTH\_BEG}}\rangle$ $\langle\text{\texttt{T\_NUMBER\_BEG}}\rangle$ 9078563412 $\langle\text{\texttt{T\_NUMBER\_END}}\rangle$ $\langle\text{\texttt{T\_EMAIL\_BEG}}\rangle$ noah@gmail.com $\langle\text{\texttt{T\_EMAIL\_END}}\rangle$ $\langle\text{\texttt{T\_NAME\_BEG}}\rangle$ William $\langle\text{\texttt{T\_NAME\_END}}\rangle$ $\langle\text{\texttt{T\_CONTACT\_BOTH\_END}}\rangle$ \\ \cline{2-4}
                
                & Contact, Name 
                & 6974028567, Benjamin
                & $\langle\text{\texttt{T\_CONTACT\_CALL\_BEG}}\rangle$ $\langle\text{\texttt{T\_NUMBER\_BEG}}\rangle$ 6974028567 $\langle\text{\texttt{T\_NUMBER\_END}}\rangle$ $\langle\text{\texttt{T\_NAME\_BEG}}\rangle$ Benjamin $\langle\text{\texttt{T\_NAME\_END}}\rangle$ $\langle\text{\texttt{T\_CONTACT\_CALL\_END}}\rangle$ \\ \cline{2-4}
                
                & Email, Name
                & alexander@gmail.com, Lucas 
                & $\langle\text{\texttt{T\_CONTACT\_MAIL\_BEG}}\rangle$ $\langle\text{\texttt{T\_EMAIL\_BEG}}\rangle$ alexander@gmail.com $\langle\text{\texttt{T\_EMAIL\_END}}\rangle$ $\langle\text{\texttt{T\_NAME\_BEG}}\rangle$ Lucas $\langle\text{\texttt{T\_NAME\_END}}\rangle$ $\langle\text{\texttt{T\_CONTACT\_MAIL\_END}}\rangle$ \\ 
                
                \midrule
                \multirow{3}{*}{CONTACT ::SHARE}
                
                & Contact, Email, Name 
                & 9067894523, henry@gmail.com, Oliver
                & Oliver's contact number is 9067894523 and email address is henry@gmail.com \\ \cline{2-4}
                
                & Contact, Name
                & 6384559067, Liam  
                & You can connect with Liam at 6384559067 \\ \cline{2-4}
                
                & Email, Name  
                & benjamin@gmail.com, Noah 
                & Noah's email address is benjamin@gmail.com \\ 
                
                \midrule
                LOCATION ::SHARE
                
                & Location
                & SMS Hospital Jaipur 
                & Right now, I am at SMS Hospital Jaipur \\
                
                \midrule
                \multirow{6}{*}{CALENDAR ::ACT}
                
                & Date, Location, Title, Name, ID, TimeStart, TimeEnd
                & Monday 29 August 2022, Saint Albert Hall Jaisalmer, Agenda Confrontation, Lucas, MOID\_1001043, 12:50, 13:50 
                & $\langle\text{\texttt{T\_MEETING\_EDIT\_BEG}}\rangle$ $\langle\text{\texttt{T\_DATE\_BEG}}\rangle$ Monday 29 August 2022 $\langle\text{\texttt{T\_DATE\_END}}\rangle$ $\langle\text{\texttt{T\_LOC\_BEG}}\rangle$ Saint Albert Hall Jaisalmer $\langle\text{\texttt{T\_LOC\_END}}\rangle$ $\langle\text{\texttt{T\_TITLE\_BEG}}\rangle$ Agenda Confrontation $\langle\text{\texttt{T\_TITLE\_END}}\rangle$ $\langle\text{\texttt{T\_NAME\_BEG}}\rangle$ LUCAS $\langle\text{\texttt{T\_NAME\_END}}\rangle$ $\langle\text{\texttt{T\_ID\_BEG}}\rangle$ MOID\_1001043 $\langle\text{\texttt{T\_ID\_END}}\rangle$ $\langle\text{\texttt{T\_TIMES\_BEG}}\rangle$ 12:50 $\langle\text{\texttt{T\_TIMES\_END}}\rangle$ $\langle\text{\texttt{T\_TIMEE\_BEG}}\rangle$ 13:50 $\langle\text{\texttt{T\_TIMEE\_END}}\rangle$ $\langle\text{\texttt{T\_MEETING\_EDIT\_END}}\rangle$ \\ \cline{2-4}
                
                & Date, Location, Title, ID, TimeStart, TimeEnd 
                & ... 
                & ... \\ \cline{2-4}
                
                & Date, Title, ID, TimeStart, TimeEnd  
                & ... 
                & ... \\ \cline{2-4}
                
                & Location, Title, Name, ID  
                & ... 
                & ... \\ \cline{2-4}
                
                & Location, Name, ID  
                & ... 
                & ... \\ \cline{2-4}
                
                & Name, ID, TimeStart  
                & ... 
                & ... \\
                
                \midrule
                \multirow{6}{*}{CALENDAR ::SHARE} 
                
                & Date, Location, Title, Name, ID, TimeStart, TimeEnd  
                & Monday 29 August 2022, Saint Albert Hall Jaisalmer, Success Strategies, William, MOID\_1001039, 17:50, 18:50 
                & The meeting MOID\_1001039 has been scheduled with William on Monday 29 August 2022 from 17:50 to 18:50 at Saint Albert Hall Jaisalmer regarding Success Strategies \\ \cline{2-4}
                
                & Date, Location, Title, ID, TimeStart, TimeEnd 
                & 11th June 2022, Board Room, Moving to Mastery, MOID\_1000259, 12:25, 14:25 
                & The meeting with ID MOID\_1000259 has been scheduled at Board Room regarding Moving to Mastery on 11th June 2022 at 12:25 to 14:25 \\ \cline{2-4}
                
                & Date, Title, ID, TimeStart, TimeEnd 
                & 21st March 2022, Moving to Mastery, MOID\_1000759, 12:25, 13:25 
                & The meeting with ID MOID\_1000759 has been scheduled regarding Moving to Mastery on 21st March 2022 at 12:25 to 13:25 \\ \cline{2-4}
                
                & Location, Title, Name, ID  
                & Saint Albert Hall, Everything Counts, William, MOID\_1001041 
                & You have a meeting MOID\_1001041 with William regarding Everything Counts at Saint Albert Hall \\ \cline{2-4}
                
                & Location, Name, ID  
                & Starbucks, Jacob, MOID\_1111041 
                & MOID\_1111041 is scheduled with Jacob at Starbucks \\ \cline{2-4}
                
                & Name, ID, TimeStart 
                & John, MOID\_1111081, 13:45 
                & You have a meeting MOID\_1001041 with John at 13:45  \\
                
                \midrule
                OCCASION ::SHARE
                
                & Name, Occasion  
                & Lucas, New Year & Wish you a very Happy New Year Lucas \\
                
                \midrule
                \bottomrule
    			
    		\end{tabular}
		\end{adjustbox}
	}
	\label{tab:customExamples}
\end{table*}

\end{document}